\documentclass[10pt,twocolumn,letterpaper]{article}

\usepackage[pagenumbers]{cvpr} 

\usepackage[table,xcdraw]{xcolor}
\usepackage{multirow}

\usepackage[accsupp]{axessibility}  

\usepackage{rotating} 
\newcommand{\rb}[1]{
  \rotatebox{90}{#1}
}

\definecolor{cvprblue}{rgb}{0.21,0.49,0.74}
\usepackage[pagebackref,breaklinks,colorlinks,allcolors=cvprblue]{hyperref}

\def\paperID{} 
\def\confName{CVPR}
\def\confYear{2026}

\usepackage{fancyhdr}
\fancypagestyle{arxivhdr}
{
   \fancyhf{}
   \setlength{\headheight}{15pt} 
\fancyfoot[C]{This paper has been accepted at the 2026 IEEE/CVF Conference on Computer Vision and Pattern Recognition (CVPR)
}
\fancyhead[C]{\footnotesize Please cite this paper as:\\
S. Mosco, D. Fusaro, A. Pretto, "Learning to Identify Out-of-Distribution Objects for 3D LiDAR Anomaly Segmentation,"\\IEEE/CVF Conference on Computer Vision and Pattern Recognition (CVPR),  2026}
}

\title{Learning to Identify Out-of-Distribution Objects\\for 3D LiDAR Anomaly Segmentation}

\author{Simone Mosco \quad
        Daniel Fusaro \quad
        Alberto Pretto \\
        University of Padova, Italy\\
{\tt\small \{moscosimon, fusarodani, alberto.pretto\}@dei.unipd.it}
}

\begin{document}
\maketitle

\thispagestyle{arxivhdr}



\begin{abstract}

Understanding the surrounding environment is fundamental in autonomous driving and robotic perception. Distinguishing between known classes and previously unseen objects is crucial in real-world environments, as done in Anomaly Segmentation. However, research in the 3D field remains limited, with most existing approaches applying post-processing techniques from 2D vision. To cover this lack, we propose a new efficient approach that directly operates in the feature space, modeling the feature distribution of inlier classes to constrain anomalous samples. Moreover, the only publicly available 3D LiDAR anomaly segmentation dataset contains simple scenarios, with few anomaly instances, and exhibits a severe domain gap due to its sensor resolution. To bridge this gap, we introduce a set of mixed real-synthetic datasets for 3D LiDAR anomaly segmentation, built upon established semantic segmentation benchmarks, with multiple out-of-distribution objects and diverse, complex environments.
Extensive experiments demonstrate that our approach achieves state-of-the-art and competitive results on the existing real-world dataset and the newly introduced mixed datasets, respectively, validating the effectiveness of our method and the utility of the proposed datasets.
Code and datasets are available at \href{https://simom0.github.io/lido-page/}{https://simom0.github.io/lido-page/}.

\end{abstract}


\section{Introduction}
\label{sec:intro}

Autonomous vehicles and robotic systems need to perceive the surrounding environment to operate effectively. LiDAR semantic segmentation~\cite{guo2020deep} is a crucial task, which enables understanding of 3D scenes through point-wise classification. Several approaches~\cite{kong2023rethinking, wu2024point, feng2024lsk3dnet} address this task by exploiting different LiDAR data representations. However, a crucial challenge in real-world scenarios is the ability to identify and segment unknown objects, not previously seen in training (\textit{i.e.}, anomalies). Most methods are trained on standard closed-set assumptions with a fixed set of classes and struggle to identify unknown objects.

\begin{figure}[t]
    \centering
    \includegraphics[width=0.99\linewidth]{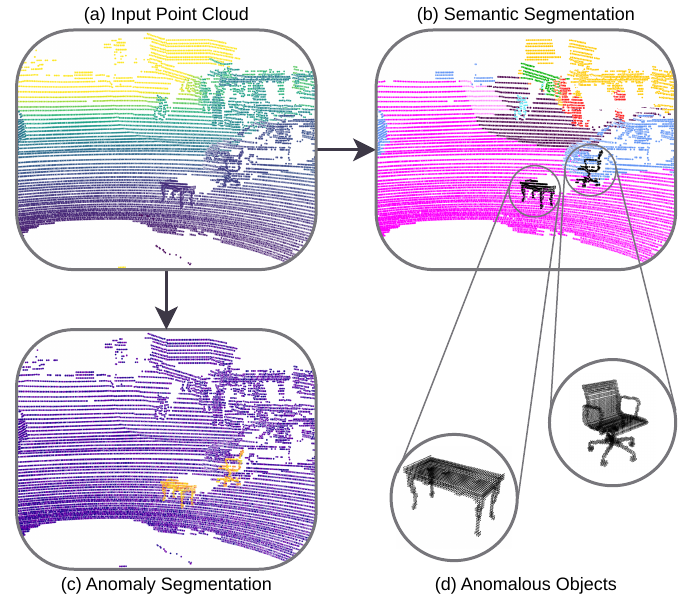}
    \caption{Overview of the LiDAR anomaly segmentation task. Given an input point cloud containing a previously unseen object (a), the purpose is to simultaneously perform semantic segmentation on known classes (b) and assign each point a score (c) indicating its probability of belonging to an anomalous object (d).}
    \label{fig:anomaly-overview}
\end{figure}

Recent approaches for anomaly detection and anomaly segmentation~\cite{delic2024outlier, nayal2023rba, sodano2024cvpr, nayal2025likelihood} achieve promising progress in these tasks, supported by the availability of several datasets~\cite{blum2021fishyscapes, chan2021segmentmeifyoucan, li2022coda}. However, they are mainly related to the image domain, while most autonomous driving datasets~\cite{semantickittidataset, nuscenes, pan2020semanticposs} rely on LiDAR sensors for robust 3D perception. Some works~\cite{wong2020identifying, chakravarthy2025lidar} address the open-set problem in 3D but struggle to segment anomalous objects or depend on unlabeled regions during training, considering them as anomalies. More recent baselines~\cite{nekrasov2025spotting} adapt simple post-processing techniques or use model ensemble~\cite{lakshminarayanan2017simple}, which is computationally expensive and slow. In contrast, our method efficiently operates in the feature space, learning the distribution of inlier classes without relying on unlabeled regions or anomaly objects during training.

Furthermore, only a few LiDAR datasets for anomaly segmentation are available. The STU dataset~\cite{nekrasov2025spotting} contains high-resolution point clouds, introducing a large domain gap with standard training data, and includes only binary anomaly masks for evaluation. Some datasets~\cite{wong2020identifying} are proprietary and thus not publicly available, while others~\cite{li2022coda} select unlabeled objects as anomalies, meaning that some still appear in the training data. To address these limitations, we construct a set of mixed real-synthetic datasets based on popular LiDAR segmentation benchmarks. We insert synthetic anomaly objects from ModelNet~\cite{wu20153d}, ensuring a domain distribution distinct from the training classes. The resulting datasets include multiple LiDAR resolutions and frequent anomaly occurrences for diverse and challenging evaluation scenarios, with semantic labels.

The main contribution of this paper is a novel approach for LiDAR anomaly segmentation that operates directly in the feature space. Our method jointly performs semantic and anomaly segmentation by learning a distribution of inlier class prototypes, enabling the detection of unseen objects without requiring anomaly samples during training. Additionally, we introduce a new set of mixed real-synthetic datasets for LiDAR anomaly segmentation, constructed by combining real LiDAR scans and synthetic out-of-distribution objects. We design a strategy to maintain realism by aligning the geometric properties and  point distribution of the inserted objects with the LiDAR beams measurement pattern and assigning consistent remission values based on reflection models. The key contributions of this work can be summarized as:

\begin{itemize}
    \item We propose a novel deep learning approach for efficient 3D LiDAR anomaly segmentation that directly operates on the feature space.
    \item We introduce a new set of mixed real-synthetic LiDAR datasets with out-of-distribution objects, based on publicly available benchmarks.
    \item We demonstrate that our approach achieves state-of-the-art and competitive performance on the real and mixed datasets, respectively.
\end{itemize}

\section{Related Work}
\label{sec:relatedwork}

\subsection{Point Cloud Semantic Segmentation}
Point cloud semantic segmentation aims to assign a class label to each 3D point.
There exist four main categories to group these approaches, namely point-based, projection-based, voxel-based, and hybrid methods.
Point-based methods directly process the raw 3D points. Pioneering works~\cite{qi2017pointnet, qi2017pointnet++} use MLPs and symmetric pooling functions to learn per-point features. Several methods~\cite{thomas2019kpconv, wang2019dynamic, wu2019pointconv} introduce point convolution,
while others rely on the attention mechanism~\cite{zhao2021point, wu2022point, wu2024point}.
Recent works~\cite{puy2023using, fusaro2024exploiting, mosco2025point} combine both 3D and 2D operations with plane projections.
Projection-based approaches project the point cloud onto a 2D surface. Several approaches~\cite{wu2018squeezeseg, wu2019squeezesegv2, xu2020squeezesegv3, cortinhal2020salsanext, milioto2019rangenet++, cheng2022cenet} operate on the range image representation, leveraging traditional 2D backbones, others~\cite{zhang2020polarnet, aksoy2020salsanet} exploit the bird-eye-view representation. Recent works~\cite{ando2023rangevit, kong2023rethinking, mosco2026revisiting} adapt attention-based backbones from the 2D field, improving results.
Voxel-based methods represent the point cloud as a grid of 3D voxels and apply 3D convolutional operations. Initial approaches focus on computational efficiency, such as designing sparse 3D convolutions~\cite{choy20194d} or LiDAR-suitable 3D representations~\cite{zhu2021cylindrical}. Some approaches~\cite{cheng2021af2s3net, lai2023spherical} integrate attention, while recent work~\cite{feng2024lsk3dnet} redesigns 3D sparse kernels for reduced complexity.
Hybrid methods combine different point cloud representations or integrate data from other sensors. Several approaches~\cite{tang2020searching, xu2021rpvnet, hou2022point} exploit different representations, while others~\cite{yan20222dpass, liu2023uniseg} leverage also RGB images. Recent trends focus on robust embedding learning~\cite{li2025rapid} and on using 2D vision foundation models~\cite{zeid2025dino}.

\subsection{Anomaly Segmentation}
Anomaly segmentation extends the task of anomaly detection by predicting if each individual pixel in an image or point in a point cloud belongs to an anomaly.
Early approaches for 2D images~\cite{hendrycks2017baseline, chan2021entropy} develop post-processing techniques and directly work on softmax activations, some use an additional class in training for anomalies~\cite{blum2021fishyscapes, mor2018confidence}, and others apply model ensembles~\cite{lakshminarayanan2017simple} to measure disagreement. A line of work~\cite{xia2020synthesize, di2021pixel, zhao2023omnial} employs generative models to resynthesize the input image and look at dissimilar areas for anomalies. Many unsupervised methods use synthetic anomaly data and train an anomaly detector~\cite{roth2022towards, tsai2022multi, liu2023diversity}. Other approaches focus on confidence and uncertainty-based predictions~\cite{gal2016dropout, delic2024outlier}, entropy computation~\cite{nayal2025likelihood}, or modeling the feature space to discriminate anomaly features from the inlier classes~\cite{sodano2024cvpr}. Recently, vision-language models based on CLIP~\cite{zhong2022regionclip, rao2022denseclip} have been applied in anomaly segmentation. Different methods~\cite{nayal2023rba, rai2023unmasking} explore the mask prediction mechanism to directly identify anomalies. Alongside, a large amount of data~\cite{blum2021fishyscapes, chan2021segmentmeifyoucan, bacchin2024sood} is available when working with 2D images.

\begin{figure*}[t]
    \centering
    \includegraphics[width=0.99\linewidth]{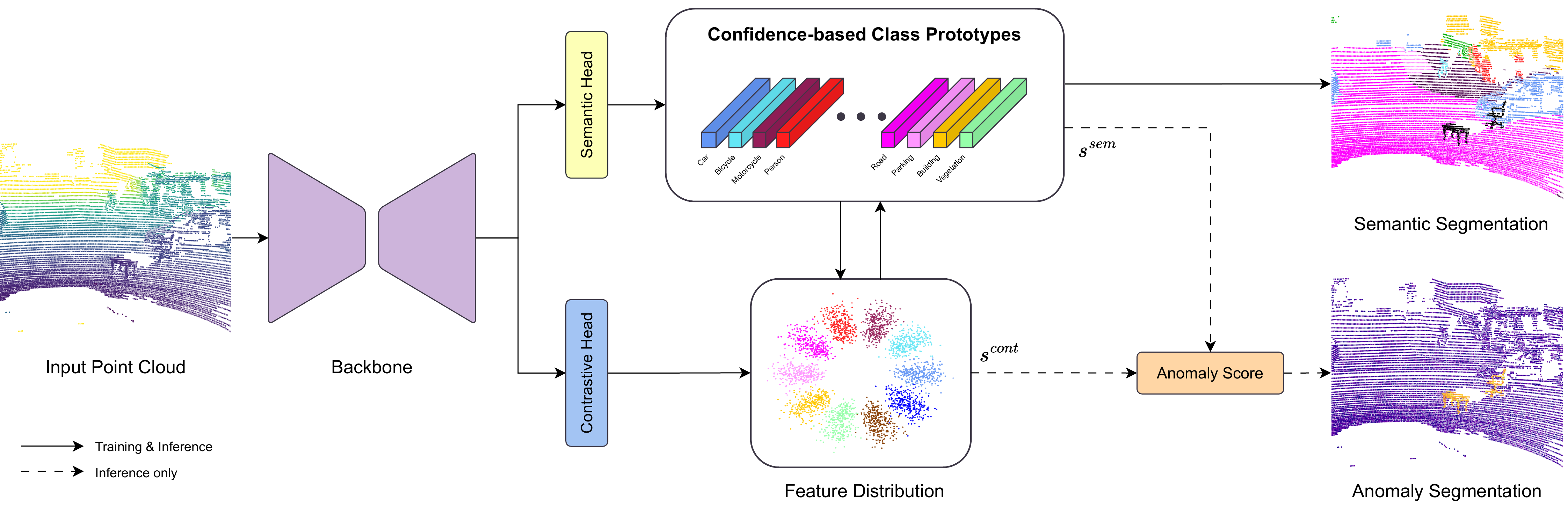}
    \caption{Overview of the proposed LIDO approach. A backbone extracts per-point features, which are then processed by two different heads. The segmentation head predicts point-wise class labels while constructing class prototypes in the feature space. The contrastive head models the distribution of inlier class features to better identify anomalies. During inference, the outputs of both heads are combined to produce the final anomaly segmentation scores.}
    \label{fig:method-overview}
\end{figure*}


LiDAR anomaly segmentation is still not well explored. Several approaches are based on 2D post-processing techniques or model ensembles~\cite{nekrasov2025spotting}, while others focus on the open-set instance segmentation task~\cite{wong2020identifying, chakravarthy2025lidar}, performing poorly on anomaly objects. Moreover, few LiDAR datasets are available for anomaly segmentation. SOD~\cite{singh2020lidar} has a few low-resolution 16-beam LiDAR data, introducing a large domain gap. TOR4D and Rare4D~\cite{wong2020identifying} contain real-world LiDAR data but are proprietary and not publicly accessible. CODA~\cite{li2022coda} is derived from existing LiDAR datasets but faces the limitation that some anomaly objects may appear in training. Recent work STU dataset~\cite{nekrasov2025spotting} provides large real-world data with anomaly annotation with high-resolution 128-beam LiDAR data.
In this work, we propose a new approach for LiDAR anomaly segmentation, inspired by~\cite{sodano2024cvpr}, along with three new datasets at different resolutions to address domain gaps in 3D anomaly segmentation.


\section{Methodology}
\label{sec:methodology}



In this section, we present the architecture of our proposed approach, \textbf{LIDO} (\textbf{L}earning to \textbf{I}dentify Out-of-\textbf{D}istribution \textbf{O}bjects), illustrated in \cref{fig:method-overview}. 
In our setting, an \emph{anomaly} denotes any point whose geometric characteristics do not correspond to any semantic class seen during training, i.e., an out-of-distribution point. LIDO is composed of a backbone to extract per-point features and two branches: the first produces semantic segmentation predictions and builds per-class prototypes, while the second directly models feature distribution to identify anomaly instances.
We adopt MinkowskiNet~\cite{choy20194d} as a backbone for feature extraction, two linear layers as heads to produce final features for each branch, and a scoring mechanism to compute anomaly predictions in inference.



\subsection{Problem Formulation}

We represent a LiDAR scan $\mathbf{X} \in \mathbb{R}^{N\times4}$ as a set of $N$ points $\mathbf{p}_n = (x_n, y_n, z_n, i_n)$, where each point consists of 3D coordinates and intensity value, respectively. For each scan, we have a corresponding set of labels $\mathbf{Y} = \{y_n\}_{n=1}^N$, where each $y_n = \{1, \dots, C\}$, encodes a class $c$ and $C$ denotes the number of inlier classes that the network sees during training. The goal of LiDAR anomaly segmentation is to produce semantic predictions $\hat{\mathbf{Y}} = \{\hat{y}_n\}_{n=1}^N$ for inlier classes, $\hat{y}_n = \{1, \dots, C\}$, and jointly assign a score $s_n \in [0, 1]$ to each point $\mathbf{p}_n$, to predict the probability of a point being an anomaly or not. We denote with $\mathbf{X}_c = \{\mathbf{p}_n \in  \mathbf{X} | y_n = c\}$ the set of points whose ground truth class is $c$, and with $\hat{\mathbf{X}}_c = \{\mathbf{p}_n \in  \mathbf{X}_c | \hat{y}_n = y_n\}$ the set of true positive points for class $c$, where the ground truth and predicted class match.

\subsection{Semantic Head}

The semantic head is responsible for generating the semantic predictions and, at the same time, constructing a robust prototype for each inlier class. We use standard practice and optimize it with a weighted cross-entropy loss:

\begin{equation}
    \mathcal{L}_\text{ce} = -\frac{1}{N} \sum_{n=1}^{N} w_c y_n \log(\sigma(f_n)),
    \label{eq:cross-entropy}
\end{equation}

\noindent
where $w_c$ corresponds to the class weight for class $c$, $y_n$ is the ground truth class for point $\mathbf{p}_n$, $\sigma$ denotes the softmax operation and $f_n$ represents the pre-softmax feature vector predicted for point $\mathbf{p}_n$.

During training, we are also interested in the construction of class prototypes to gather all points belonging to a certain class closer in the feature space. To obtain this, we consider pre-softmax features of all true positives for each class and accumulate them. We extract a confidence value $\kappa_\mathbf{p}$ for each point by taking the maximum component of its pre-softmax feature vector $\kappa_\mathbf{p} = \text{max}(f_\mathbf{p})$ and compute a confidence-based prototype (CP) for each class $c$, $\text{CP} = \{ \text{CP}_1, \dots, \text{CP}_c \}$: 

\begin{equation}
    \text{CP}_c = \frac{\sum_{\mathbf{p} \in \hat{\mathbf{X}}_c} \kappa_{\mathbf{p}} f_\mathbf{p}}{\sum_{\mathbf{p} \in \hat{\mathbf{X}}_c} \kappa_{\mathbf{p}}},
    \label{eq:confidence-prototype}
\end{equation}

\noindent
where $f_\mathbf{p}$ are the pre-softmax features of point $\mathbf{p}$.

At the beginning of each epoch $e$, we use the confidence-based prototypes computed in the previous epoch $\text{CP}_c^{e-1}$ to guide the network into producing feature vectors for points with ground truth class $c$, close to $\text{CP}_c^{e-1}$. We achieve this by introducing a prototype-based cosine embedding loss that enforces proximity between each feature and its corresponding prototype:


\begin{equation}
    \mathcal{L}_\text{prot} = \frac{1}{N} \sum_{c \in C} \sum_{\mathbf{p} \in \mathbf{X}_c} \left( 1 - \left< \text{CP}_c^{e-1}, f_\mathbf{p} \right> \right).
    \label{eq:prototype-loss}
\end{equation}

In the first epoch, this loss is not active as there are no accumulated prototypes yet. Overall, the semantic head is optimized with a weighted sum of the above-mentioned losses, combined with the Lovasz Loss~\cite{berman2018lovasz} $\mathcal{L}_\text{lovasz}$:

\begin{equation}
    \mathcal{L}_{\text{shead}} = \lambda_1 \mathcal{L}_{\text{ce}} + \lambda_2 \mathcal{L}_{\text{lovasz}} + \lambda_3 \mathcal{L}_{\text{prot}}.
    \label{eq:semantic-loss}
\end{equation}

\subsection{Contrastive Head}

The aim of the contrastive head is to directly identify points belonging to anomalies by learning discriminative and distinguished per-class prototypes, and modeling their distribution in the feature space, similar to~\cite{sodano2024cvpr}.
To achieve this, we adopt both the contrastive loss~\cite{chen2020simple} and the objectosphere loss~\cite{dhamija2018reducing}.

We denote with $f_\mathbf{p}'$ the pre-softmax features produced by the contrastive head for point $\mathbf{p}_n$ and compute the mean feature vector $\bar{f}_c$ for each class $c$ as:

\begin{equation}
    \bar{f}_c = \frac{1}{|\mathbf{X}_c|} \sum_{\mathbf{p} \in \mathbf{X}_c} f_\mathbf{p}',
\end{equation}

\noindent
considering all the points whose ground truth label is $c$, with $|\mathbf{X}_c|$ the cardinality of $\mathbf{X}_c$. Then, we employ the contrastive loss $\mathcal{L}_\text{cont}$ to align the mean feature vector $\bar{f}_c$ to the normalized confidence-based prototype $\text{CP}_c^{e-1}$ for the corresponding class $c$, at the previous epoch, and at the same time push them away from the feature vectors of the other classes:

\begin{equation}
    \mathcal{L}_\text{cont} = - \sum_{c \in C} \log \frac{\exp (  \left < \bar{f}_c, \text{CP}_c^{e-1} \right > / \tau ) }{\sum_{i=1}^C \exp (  \left < \bar{f}_c, \text{CP}_c^{e-1} \right > / \tau )},
    \label{eq:contrastive-loss}
\end{equation}

\noindent
where $\tau$ is a temperature parameter. In addition, we use the objectosphere loss $\mathcal{L}_{obj}$ over each point $\mathbf{p} \in \mathbf{X}$, denoted as:

\begin{equation}
    \mathcal{L}_\text{obj} =
    \begin{cases}
        \text{max}(r - \|f_\mathbf{p}'\|^2, 0) & \text{if } \mathbf{p} \in \mathcal{D}_{in} \\
        \|f_\mathbf{p}'\|^2 & \text{otherwise}
    \end{cases},
    \label{eq:obj-loss} 
\end{equation}

\noindent
where $f_\mathbf{p}'$ is the feature vector of a point $\mathbf{p}$ belonging to the set $\mathcal{D}_{in}$ of inlier classes, and $r$ is a fixed threshold. Different from~\cite{sodano2024cvpr}, we do not use unlabeled or void regions in training to learn anomaly features; thus, the goal of this loss is to push feature vectors of inlier classes far from the center of the $C$-dimensional hypersphere of radius $r$.

The contrastive head is then optimized with a weighted sum of these two losses as follows:

\begin{equation}
    \mathcal{L}_{\text{chead}} = \lambda_4 \mathcal{L}_{\text{cont}} + \lambda_5 \mathcal{L}_{\text{obj}}.
    \label{eq:conthead-loss}
\end{equation}

\subsection{Inference}
\label{sec:inference}

In order to obtain both semantic segmentation and anomaly predictions, we combine the outputs of the two heads. The semantic head provides standard semantic segmentation predictions through cosine similarity among confidence-based prototypes $\text{CP}_c$ for class $c$, accumulated during training, and per-point features $f_n$, for point $\mathbf{p}_n$:

\begin{equation}
    \text{sim}_{n,c} = \left< f_n, \text{CP}_c \right>.
    \label{eq:cosine-similarity}
\end{equation}

The predicted inlier semantic class is obtained considering the maximum similarity values among all classes:

\begin{equation}
    \hat{y}_n = \text{argmax}_c(\text{sim}_{n,c}).
    \label{eq:semantic-predictions}
\end{equation}

The semantic head also produces a score that estimates if a point $\mathbf{p}_n$ belongs to an anomaly object, combining both the cosine distance and the entropy of the softmax values. First, we denote as $s_n^{cos}$ the score corresponding to the maximum cosine distance:

\begin{equation}
    s_n^{cos} = 1 - \text{max}_c(\text{sim}_{n,c}),
    \label{eq:score-cosine-distance}
\end{equation}

\noindent
where an high score considers the point $\mathbf{p}_n$ as anomaly. Second, we compute an entropy-based score $s_n^{ent}$ using the normalized Shannon entropy~\cite{shannon1948entropy} as:

\begin{equation}
    s_n^{ent} = - \frac{1}{\log C} \sum_{c \in C} p_{n,c}\log (p_{n,c}),
    \label{eq:score-entropy}
\end{equation}

\noindent
where $p_{n,c}$ is the softmax probability of point $\mathbf{p}_n$ belonging to class $c$, obtained from $p_n = \sigma(f_n)$, with $\sigma$ that denotes the softmax operation. Higher entropy values indicates the network uncertainty on assigning a point to a specific class, meaning that it may belong to an anomaly.

Then, we combine these two scores and obtain a point-wise score $s_n^{sem} = s_n^{cos} \cdot s_n^{ent}$ and normalize it with respect to the maximum value, to ensure values in range $[0, 1]$.

The contrastive head produces another anomaly prediction scores, based on the principle of the objectosphere loss~\cite{dhamija2018reducing}. We consider anomaly all the points whose feature vectors have norm below a certain threshold $r$, as specified in~\cref{eq:obj-loss}. Thus, we define a point-wise score $s_n^{cont}$ for point $\mathbf{p}_n$ as:

\begin{equation}
    s_n^{cont} = \text{max}\left ( 0, \left ( 1 - \frac{\| f_n' \|^2}{r} \right ) \right ).
    \label{eq:score-contrastive}
\end{equation}

This score is 1 when the feature vector has norm equals to 0, while 0 when the norm is greater that the threshold $r$.

Finally, we obtain a per-point score for belonging to an anomaly, fusing the two scores from the heads as:

\begin{equation}
    s_n = \frac{1}{2} \left (s_n^{sem} + s_n^{cont} \right ).
    \label{eq:score-final}
\end{equation}


\section{Out-of-Distribution Datasets}
\label{sec:datasets}

\begin{table}[t]
    \caption{Comparison of publicly available LiDAR datasets for Anomaly Segmentation. BB indicates bounding box, SM semantic masks, and BM binary masks. Size refers to the number of scans in the dataset, \#OoD Instances denotes the total number of anomaly objects.}
    \label{tab:dataset-comparison}
    \centering
    \resizebox{0.47\textwidth}{!}{%
    \begin{tabular}{lcccc}
        \toprule
        Dataset & \#Beams & Size & Labels & \#OoD Instances \\ 
        \midrule
        CODA-KITTI~\cite{li2022coda} & 64 & 309 & BB & 399 \\
        CODA-nuScenes~\cite{li2022coda} & 32 & 134 & BB & 1125 \\
        CODA-ONCE~\cite{li2022coda} & 40 & 1057 & BB & 4413 \\
        SOD~\cite{singh2020lidar} & 16 & 460/530 & SM & - \\
        STU~\cite{nekrasov2025spotting} & 128 & 8022/1960 & BM & 1965/- \\
        \midrule
        \rowcolor{green!15}
        \textbf{nuScenes-OoD (Ours)} & 32 & 6019 & SM & 2398/7268 \\
        \rowcolor{green!15}
        \textbf{SemanticPOSS-OoD (Ours)} & 40 & 500 & SM & 196/586 \\
        \rowcolor{green!15}
        \textbf{SemanticKITTI-OoD (Ours)} & 64 & 4071 & SM & 1634/4894 \\
        \bottomrule
    \end{tabular}}
\end{table}



In this section, we introduce our mixed real-synthetic Out-of-Distribution (OoD) datasets for 3D LiDAR anomaly segmentation. The datasets (see~\cref{tab:dataset-comparison}) are constructed from three autonomous driving benchmarks with different LiDAR sensor resolutions, complementary to the only available real-world dataset, STU~\cite{nekrasov2025spotting}. We use ModelNet~\cite{wu20153d} as a source for synthetic anomaly objects, filtering its models to avoid overlap with categories and objects present in real-world LiDAR datasets (see Supplementary Material).
To ensure realism, we also introduce a protocol for inserting synthetic objects into real LiDAR scans, manipulating point distributions, intensity values, and aligning them to the LiDAR sensor geometry in a beam-like format. The proposed OoD datasets are as follows:



\textbf{nuScenes-OoD} is constructed from the official validation set of nuScenes~\cite{nuscenes} dataset, comprising 6019 scans collected with a 32-beam LiDAR sensor.

\textbf{SemanticPOSS-OoD} is derived from the 500 samples in the validation sequence of SemanticPOSS~\cite{pan2020semanticposs}, acquired with a 40-beam LiDAR sensor.

\textbf{SemanticKITTI-OoD} is based on the 4071 scans in the validation set of SemanticKITTI~\cite{semantickittidataset}, generated with a 64-beam LiDAR sensor.

Each dataset contains two different versions, a \textit{single} and a \textit{multi} split, comprising respectively a single anomaly object or multiple ones in each single scan. To better resemble the real-world environment, where anomalies are not so frequent, but at the same time provide a strong evaluation setup, about 40\% of the scans in the \textit{single} split contain anomalies, while 60\% for the \textit{multi} split. Moreover, to set a different level of difficulty, anomaly objects in the \textit{single} split are inserted only on the road, while in the \textit{multi} split, also on other surfaces, as parking areas or sidewalks. 

\subsection{Protocol}



Consider a real LiDAR point cloud $\mathbf{P} \in \mathbb
{R}^{N\times 4}$ with $N$ points, we randomly sample a model, from the filtered ModelNet dataset, $\mathbf{O} \in \mathbb{R}^{M\times 4}$, with $M$ points and inject it into the LiDAR scan, resulting in a combined point cloud $\mathbf{S} = [\mathbf{P}, \mathbf{O}] \in \mathbb{R}^{(N+M)\times 4}$. For the insertion, we consider only planar surfaces: solely the road for the \textit{single} split, and other available planar surfaces (e.g., road, sidewalks, parking areas, etc.) for the \textit{multi} split. Standard augmentation techniques as rotation and scaling, are applied to the object points, ensuring a coherent size ratio with respect to the real-world scan.
Points in $\mathbf{S}$ are then projected into a range image representation~\cite{milioto2019rangenet++}, and re-projected in the 3D space, resulting in a point cloud $\mathbf{S}' = [\mathbf{P}', \mathbf{O}']$, with $\mathbf{P}'$ and $\mathbf{O}'$ the filtered scan and object point clouds, respectively.
The re-projection operation allows us to recover information from original points potentially occluded during object insertion and, at the same time, geometrically align the object points to the LiDAR spatial property, sampling points from the inserted object in the LiDAR beams format.
Since ModelNet objects are not provided with intensity values, and earlier steps used temporary intensity values, we compute them using the Lambertian reflectance model~\cite{oren1995generalization}. Different from simulators as CARLA~\cite{dosovitskiy2017carla}, which compute intensity in a simplified way, based solely on the distance from the sensor, our approach relies on physical model properties to better match the real-world behavior. The intensity $i$ is computed as:

\begin{equation}
    i = \frac{\rho \cdot \text{max}(0, - \left < \mathbf{n}, \mathbf{r} \right >)}{d^2},
    \label{eq:intesity-formula}
\end{equation}

\noindent
where $\rho$ is the reflectivity value of the object, $\mathbf{n}$ is the surface normal at the point (estimated from its neighbors), $\mathbf{r}$ is the LiDAR beam direction towards the point, and $d$ is the distance from the sensor. When the normal faces towards the sensor, the intensity value is higher, while it falls to 0 if the normal faces sideways or away. The reflectivity value $\rho$ is an inherent property of the model's material, which measures how much light a surface reflects. We assign these values based on real object measurements (see Supplementary Material). To increase consistency with the target scan, we normalize the intensity values of the object with respect to the average intensity of the scan in which it is inserted and add a small per-point Gaussian perturbation as noise to better match the real-world properties.
Finally, we obtain a mixed real-synthetic LiDAR scan with one or more injected ModelNet models representing anomalies, by just fusing points from the two refined point clouds and updating labels accordingly.


\section{Experiments}
\label{sec:experiments}

\begin{table*}[t]
    \caption{Anomaly Segmentation performance on STU dataset.}
    \label{tab:exp-stu}
    \centering
    \resizebox{0.99\linewidth}{!}{
    \begin{tabular}{lcccccc}
        \toprule
         & \multicolumn{3}{c}{Validation Set} & \multicolumn{3}{c}{Test Set} \\
        Method & AUROC [\%] $\uparrow$ & FPR@95 [\%] $\downarrow$ & AP [\%] $\uparrow$ & AUROC [\%] $\uparrow$ & FPR@95 [\%] $\downarrow$ & AP [\%] $\uparrow$ \\
         \midrule
        Mask4Former3D + MC Dropout~\cite{srivastava2014dropout} & 65.76 & 79.82 & 0.17 & 61.51 & 82.37 & 0.11 \\
        Mask4Former3D + RbA~\cite{nayal2023rba} & 73.00 & 100.0 & 1.64 & 66.38 & 100.0 & 0.81 \\
        Mask4Former3D + Max Logit~\cite{hendrycks2017baseline} & 87.27 & 68.76 & 2.02 & 84.53 & 81.49 & 0.95 \\
        Mask4Former3D + Void Classifier~\cite{blum2021fishyscapes} & 89.77 & 79.50 & 2.62 & 85.99 & 78.60 & 3.92 \\
        Mask4Former3D + Deep Ensemble~\cite{lakshminarayanan2017simple} & \underline{90.93} & \underline{37.34} & \underline{6.94} & \underline{86.74} & \underline{58.05} & \underline{5.17} \\
        \rowcolor{green!15}
        \textbf{LIDO (ours)} & \textbf{95.05} & \textbf{34.86} & \textbf{27.53} & \textbf{93.67} & \textbf{34.29} & \textbf{14.99} \\
         \bottomrule
    \end{tabular}}
\end{table*}

\begin{table*}[t]
    \caption{Anomaly Segmentation performance on SemanticPOSS-OoD dataset.}
    \label{tab:exp-poss}
    \centering
    \resizebox{0.99\linewidth}{!}{
    \begin{tabular}{lcccccc}
        \toprule
         & \multicolumn{3}{c}{Single} & \multicolumn{3}{c}{Multi} \\
        Method & AUROC [\%] $\uparrow$ & FPR@95 [\%] $\downarrow$ & AP [\%] $\uparrow$ & AUROC [\%] $\uparrow$ & FPR@95 [\%] $\downarrow$ & AP [\%] $\uparrow$ \\
         \midrule
        Mask4Former3D + RbA~\cite{nayal2023rba} & 49.09 & 100.0 & 0.13 & 50.42 & 100.0 & 0.26 \\
        Mask4Former3D + Max Logit~\cite{hendrycks2017baseline} & 61.05 & 91.85 & 0.19 & 63.77 & 88.74 & 0.40 \\
        Mask4Former3D + Deep Ensemble~\cite{lakshminarayanan2017simple} & \underline{85.86} & \underline{58.12} & \underline{0.82} & \underline{85.83} & \underline{54.20} & \underline{1.21} \\
        \rowcolor{green!15}
        \textbf{LIDO (ours)} & \textbf{91.51} & \textbf{45.10} & \textbf{3.97} & \textbf{90.84} & \textbf{44.74} & \textbf{5.92} \\
         \bottomrule
    \end{tabular}}
\end{table*}

The experimental section is designed to support our claims that: (i) our approach effectively works on the feature space to address the 3D LiDAR anomaly segmentation task; (ii) the proposed mixed real-synthetic datasets provide a valuable and challenging benchmark for this problem; (iii) the proposed approach achieves competitive performance on both real and mixed datasets.   

\subsection{Implementation Details}

\noindent
\textbf{Evaluation Setup.} 
We evaluate our approach on four datasets: the STU~\cite{nekrasov2025spotting} dataset and the three introduced mixed real-synthetic OoD datasets.
For STU, we train our model solely on the SemanticKITTI~\cite{semantickittidataset} training split.
We use the official STU split, with 19 sequences for validation and 51 for test.
For our mixed real-synthetic OoD datasets, we follow standard training procedures used in LiDAR semantic segmentation. Models are trained on the corresponding base dataset, SemanticKITTI~\cite{semantickittidataset}, SemanticPOSS~\cite{pan2020semanticposs}, nuScenes~\cite{nuscenes}, and evaluated on our modified validation split, which includes out-of-distribution objects (~\cref{sec:datasets}).

\noindent
\textbf{Training Setup.} We train our model from scratch on a single NVIDIA A40 GPU for 64 epochs, with a batch size of 4 for all datasets. We use an SGD optimizer and a cosine annealing scheduler with linear warm-up. The learning rate increases to $2.4 \times 10^{-1}$ over the first 5 epochs and then decreases to $1\times 10^{-2}$, with weight decay $1 \times 10^{-4}$. We set $r = 5.0$, $\tau = 0.1$ and loss weights $\lambda_1 = 1.0$, $\lambda_2 = 1.5$, $\lambda_3 = 0.1$, $\lambda_4 = 0.5$, and $\lambda_5 = 0.5$.
During training, we use standard augmentation techniques as rotation, flip, and scale. We do not use model ensembles nor test-time augmentations.

\noindent
\textbf{Metrics.} Following~\cite{nekrasov2025spotting}, we adopt common metrics~\cite{blum2021fishyscapes} for point-level evaluation on anomaly segmentation as Average Precision (AP), False Positive Rate at $95\%$ True-Positive Rate (FPR@95), and Area Under the Receiver Operating characteristic Curve (AUROC). For semantic segmentation, we use the standard mean Intersection over Union (mIoU).

\begin{table*}[t]
    \caption{Anomaly Segmentation performance on SemanticKITTI-OoD dataset.}
    \label{tab:exp-kitti}
    \centering
    \resizebox{0.99\linewidth}{!}{
    \begin{tabular}{lcccccc}
        \toprule
         & \multicolumn{3}{c}{Single} & \multicolumn{3}{c}{Multi} \\
        Method & AUROC [\%] $\uparrow$ & FPR@95 [\%] $\downarrow$ & AP [\%] $\uparrow$ & AUROC [\%] $\uparrow$ & FPR@95 [\%] $\downarrow$ & AP [\%] $\uparrow$ \\
         \midrule
        Mask4Former3D + RbA~\cite{nayal2023rba} & 59.68 & 100.0 & 4.56 & 71.39 & 100.0 & \underline{13.63} \\
        Mask4Former3D + Max Logit~\cite{hendrycks2017baseline} & 76.49 & 99.63 & 4.84 & 84.05 & 99.32 & \textbf{13.84} \\
        Mask4Former3D + Deep Ensemble~\cite{lakshminarayanan2017simple} & \underline{92.87} & \textbf{27.69} & \underline{6.20} & \textbf{92.19} & \textbf{28.64} & 12.04 \\
        \rowcolor{green!15}
        \textbf{LIDO (ours)} & \textbf{93.36} & \underline{31.19} & \textbf{10.60} & \underline{89.89} & \underline{39.04} & 9.42 \\
         \bottomrule
    \end{tabular}}
\end{table*}

\begin{table*}[t]
    \caption{Anomaly Segmentation performance on nuScenes-OoD dataset}
    \label{tab:exp-nuscenes}
    \centering
    \resizebox{0.99\linewidth}{!}{
    \begin{tabular}{lcccccc}
        \toprule
         & \multicolumn{3}{c}{Single} & \multicolumn{3}{c}{Multi} \\
        Method & AUROC [\%] $\uparrow$ & FPR@95 [\%] $\downarrow$ & AP [\%] $\uparrow$ & AUROC [\%] $\uparrow$ & FPR@95 [\%] $\downarrow$ & AP [\%] $\uparrow$ \\
         \midrule
        Mask4Former3D + RbA~\cite{nayal2023rba} & 32.65 & 100.0 & 3.56 & 71.60 & 98.68 & 7.95 \\
        Mask4Former3D + Max Logit~\cite{hendrycks2017baseline} & 66.17 & 98.60 & 3.54 & 43.54 & 100.0 & 8.98 \\
        Mask4Former3D + Deep Ensemble~\cite{lakshminarayanan2017simple} & \textbf{91.79} & \textbf{34.59} & \textbf{18.34} & \textbf{88.62} & \textbf{43.63} & \textbf{23.87} \\
        \rowcolor{green!15}
        \textbf{LIDO (ours)} & \underline{89.33} & \underline{39.70} & \underline{6.79} & \underline{87.25} & \underline{44.51} & \underline{10.53} \\
         \bottomrule
    \end{tabular}}
\end{table*}

\subsection{3D LiDAR Anomaly Segmentation}

The first experiment shows that our approach achieves state-of-the-art results on STU validation and test sets, as reported in~\cref{tab:exp-stu}. Our method surpasses all other approaches across all metrics, including ensemble-based~\cite{lakshminarayanan2017simple} approaches ($\textbf{+9.82\%}$ AP). Despite the significant domain gap, the proposed method effectively models the feature distribution of inlier classes, improving the segmentation of anomaly objects, showing high AP and robust FPR metrics. We attribute this to the discriminative class-wise features learned during training, which mitigate the domain gap.

The second set of experiments highlights the value of the proposed mixed real-synthetic OoD datasets and shows that our method achieves competitive results on these benchmarks as well. Given the recent popularity of 3D LiDAR anomaly segmentation task, and the limited number of available approaches, we retrained several baselines from~\cite{nekrasov2025spotting} built upon~\cite{yilmaz2024mask4former}, whose implementations are publicly available, for a fair comparison. Extended experiments and further baseline comparisons are provided in the Supplementary Material.

Our approach significantly surpasses all other methods on both SemanticPOSS-OoD splits (~\cref{tab:exp-poss}), further demonstrating the effectiveness of modeling the feature space to distinguish between inlier and out-of-distribution classes. This benchmark, however, proves to be particularly challenging and the performance remains bounded, likely due to the large number of sparse instances in SemanticPOSS, which can be misinterpreted as anomalies during inference. The lower scan resolution also limits the number of features available for training, reducing overall performance.

\begin{figure*}[t]
    \centering
    \includegraphics[width=0.99\linewidth]{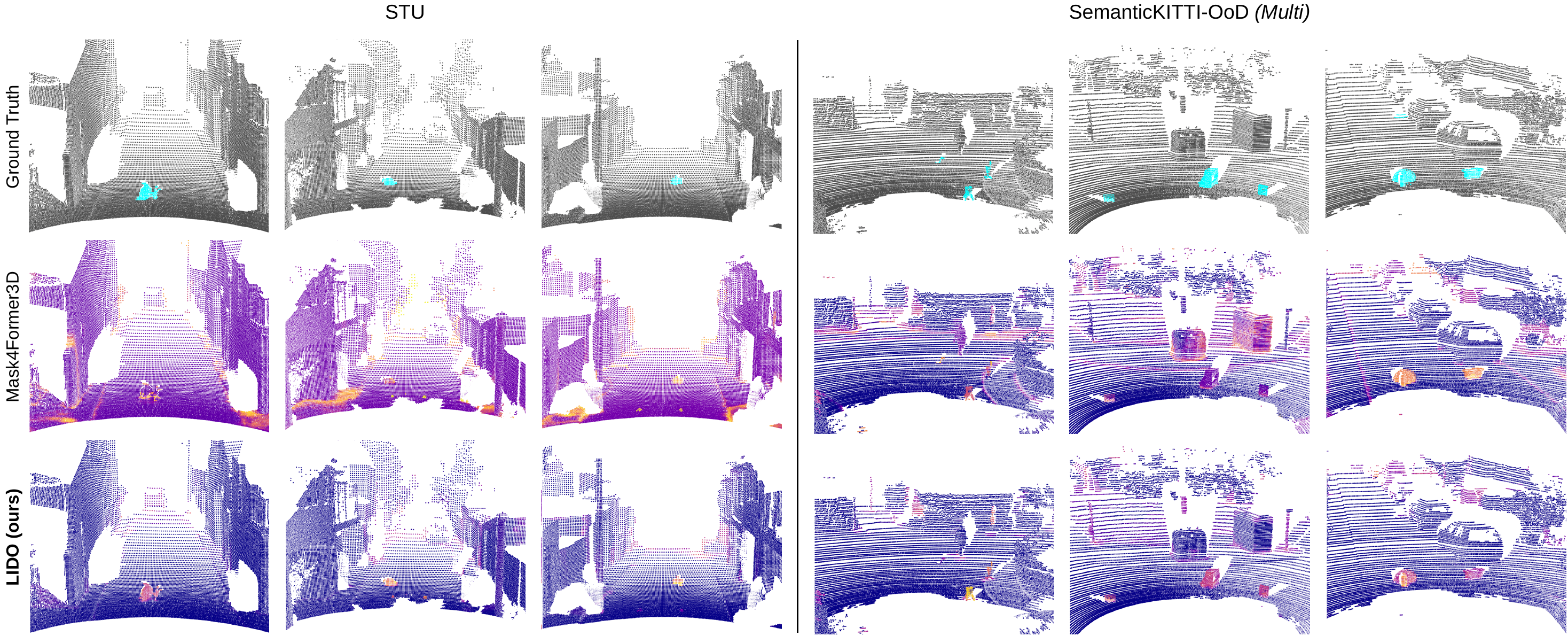}
    \caption{Anomaly Segmentation results on STU and our proposed SemanticKITTI-OoD dataset (Multi split). Ground truth anomaly objects are shown in \textcolor{cyan}{cyan}. Predicted anomaly scores range from \textcolor{blue}{blue} to \textcolor{orange}{orange}, indicating increasing probability of a point being anomalous.}
    \label{fig:qualitative-results}
\end{figure*}

~\Cref{tab:exp-kitti} reports the results on SemanticKITTI-OoD. Our method achieves state-of-the-art performance on the \textit{simple} split and competitive results on the \textit{multi} split, slightly behind the baselines from~\cite{nekrasov2025spotting}. However, these approaches exhibit a high false positive rate (FPR), see~\cref{fig:qualitative-results}, meaning that they tend to predict most points as anomalies. Moreover, the deep ensemble method, while effective, is computationally expensive and resource demanding compared to our method (see~\cref{tab:ablation-runtime}).

We report anomaly segmentation results on nuScenes-OoD in~\cref{tab:exp-nuscenes}, where our approach achieves competitive performance despite the lower resolution of the LiDAR sensor. While it does not surpass the ensemble-based approach, it is important to note that these methods require significantly higher memory and computational resources (~\cref{tab:ablation-runtime}), leading to slower inference, whereas our approach remains lightweight and efficient. The high FPR values indicate a general model uncertainty on this challenging benchmark, partially explaining the advantage of model ensembles. These findings mark the effectiveness and  practicality of the proposed feature-based approach while also suggesting room for improvement.

Overall, our approach improves upon previous methods, although metrics remain bounded, in particular, the average precision (AP). This limitation is mainly caused by the significant class imbalance inherent in LiDAR datasets~\cite{xu2021rpvnet} and the uncertainty issues observed in popular LiDAR semantic segmentation models~\cite{kong2025calib3d}. Most models often misclassify similar surfaces, classify road objects as inlier classes~\cite{hsu2020generalized}, or struggle with sparse distant points~\cite{li2025rapid}, leading to false positives. Interestingly, the higher AP values on the \textit{multi} splits may be explained by the greater number of anomaly points, which, combined with model uncertainty, tends to increase precision estimates.

We report standard semantic segmentation results in~\cref{tab:exp-semantic}. The proposed approach, with the additional losses for anomaly segmentation, maintains competitive semantic segmentation performance, with slightly lower results compared to a baseline trained in the standard setting. The only exception is nuScenes-OoD, where the lower LiDAR resolution and fewer points per scan reduce effectiveness and class prototype building, reflecting the performance on the anomaly segmentation task. See Supplementary Material for detailed class-wise segmentation metrics.  

\Cref{tab:ablation-runtime} reports computational complexity and runtime of different models. The results support our claim that the proposed approach is significantly more efficient and memory-friendly than ensemble models, achieving real-time performance ($<100$ ms) and a general balanced trade-off between accuracy and computational cost.

\begin{table}[t]
    \caption{Model complexity and runtime comparison on nuScenes-OoD dataset. Tested on a NVIDIA A40 GPU.}
    \label{tab:ablation-runtime}
    \centering
    \resizebox{0.47\textwidth}{!}{%
    \begin{tabular}{lccc}
        \toprule
        Method & Params (M) & Runtime (ms) & Memory (GB) \\
        \midrule
        Mask4Former3D~\cite{yilmaz2024mask4former} & 39.6 & 168 & 1.8 \\
        Deep Ensemble (Sequential)~\cite{lakshminarayanan2017simple} & 118.8 & 861 & 1.9 \\
        Deep Ensemble (Parallel)~\cite{lakshminarayanan2017simple} & 118.8 & 287 & 5.7 \\
        \rowcolor{green!15}
        \textbf{LIDO (ours)} & \textbf{21.7} & \textbf{38} & \textbf{0.6} \\
        \bottomrule
    \end{tabular}}
\end{table}

\begin{table}[t]
    \caption{Semantic segmentation results (mIoU, \%) of standard baseline and our proposed approach. S and M denote the single and multi splits of our OoD datasets.}
    \label{tab:exp-semantic}
    \centering
    \resizebox{0.47\textwidth}{!}{
    \begin{tabular}{lccccccc}
        \toprule
         & \multicolumn{2}{c}{KITTI-OoD} & \multicolumn{2}{c}{POSS-OoD} & \multicolumn{2}{c}{nuScenes-OoD} & STU \\
         & S & M & S & M & S & M &  \\ 
        \midrule
        Standard & \textbf{64.99} & \textbf{64.84} & \textbf{57.07} & \textbf{57.05} & \textbf{72.75} & \textbf{72.62} & \textbf{36.75} \\
        \textbf{LIDO (ours)} & 61.34 & 61.19 & 55.63 & 55.58 & 60.61 & 60.44 & 35.14 \\
        \bottomrule
    \end{tabular}}
\end{table}

\subsection{Ablation Study}

Finally, we conduct an ablation study to assess the contribution of each component in the proposed LiDAR anomaly segmentation pipeline. Starting from a baseline trained for standard LiDAR semantic segmentation, we analyze the effect of the additional loss functions introduced in the semantic and contrastive heads, together with the impact of various post-processing techniques. In~\Cref{tab:ablation-pipeline} we see that simple thresholding (A) leads to poor results, while adding the prototype loss slightly improves the results, both with thresholding (B) and cosine distance (C), though the low AP and FPR values suggest that most scores are close to zero (\textit{i.e.}, few anomalies detected). A significant improvement comes when combining prototype loss and contrastive loss, particularly when using the semantic head score (F), which benefits from incorporating entropy of~\cref{eq:score-entropy} in the score computation. Even thresholding (D) and cosine distance (E) confirm the benefit of per-class contrastive learning, pushing features apart in embedding space. The addition of the objectosphere loss yields further improvements when using the  corresponding scores (G and H), although the latter shows a higher false positive rate, indicating that relying only on the features norm is suboptimal. The combined score (I) achieves the best overall results, demonstrating the complementarity of the proposed losses.

\begin{table}[t]
    \caption{Ablation study of the LiDAR anomaly segmentation pipeline on STU validation set. $\mathcal{L}_\text{prot}$ refers to the prototype loss in~\cref{eq:prototype-loss}, $\mathcal{L}_\text{cont}$ to the contrastive loss of~\cref{eq:contrastive-loss}, and $\mathcal{L}_\text{obj}$ to  the objectosphere loss in~\cref{eq:obj-loss}. "INF" indicates the inference technique: "ML" for max logit, "C" for cosine distance, "SH" for the semantic head score, "CH" for the contrastive head score, and $s_n$ for the proposed score described in~\cref{sec:inference}.}
    \label{tab:ablation-pipeline}
    \centering
    \resizebox{0.47\textwidth}{!}{%
    \begin{tabular}{lccccccc}
        \toprule
         & $\mathcal{L}_\text{prot}$ & $\mathcal{L}_\text{cont}$ & $\mathcal{L}_\text{obj}$ & INF & AUROC $\uparrow$ & FPR@95 $\downarrow$ & AP $\uparrow$ \\
        \midrule
        (A) &  &  &  & ML & 90.92 & 43.49 & 0.97 \\
        (B) & \checkmark &  &  & ML & 92.64 & 26.60 & 1.05 \\
        (C) & \checkmark &  &  & C & 93.61 & \textbf{24.84} & 1.89 \\
        \midrule
        (D) & \checkmark & \checkmark &  & ML & 93.13 & 26.73 & 2.54 \\
        (E) & \checkmark & \checkmark &  & C & 91.71 & 34.91 & 4.86 \\
        (F) & \checkmark & \checkmark &  & SH & 95.04 & 26.71 & 12.88 \\
        \midrule
        (G) & \checkmark & \checkmark & \checkmark & SH & 92.68 & 36.57 & 14.82 \\
        (H) & \checkmark & \checkmark & \checkmark & CH & 88.16 & 100.0 & 16.67 \\
        (I) & \checkmark & \checkmark & \checkmark & $s_n$ & \textbf{95.05} & 34.86 & \textbf{27.53} \\
        \bottomrule 
    \end{tabular}}
\end{table}


\section{Conclusion}
\label{sec:conclusion}

In this paper, we introduced a novel approach for LiDAR anomaly segmentation that operates directly in the feature space, modeling the distribution of inlier class representations to identify out-of-distribution objects. Our approach combines prototype, contrastive, and objectosphere losses to constrain anomaly features in the embedding space.
In addition, we presented a new set of mixed real-synthetic out-of-distribution datasets, constructed from popular autonomous driving benchmarks with geometrically aligned synthetic anomalies, addressing the scarcity of datasets for this task.
Extensive experiments demonstrated the effectiveness of our approach, which achieves state-of-the-art and competitive results on both real and mixed datasets.


\noindent
\textbf{Future Work.} Despite the encouraging results, there is room for improvement. We plan to extend our approach to cross-domain tasks~\cite{saltori2022cosmix, li2025dpgla} to enhance generalization and to further investigate the impact of uncertainty~\cite{kong2025calib3d, miandashti2024calibrated} in LiDAR semantic segmentation to improve reliability and robustness of anomaly predictions.

{
    \small
    \bibliographystyle{ieeenat_fullname}
    \bibliography{main}
}


\clearpage
\setcounter{page}{1}
\maketitlesupplementary
\appendix


\section{Further Details on OoD Datasets}


In this section, we provide additional details about the proposed mixed real-synthetic OoD datasets, including the construction process, the selected ModelNet objects inserted into the scans, the insertion protocol, and the technique used to align point distribution with the LiDAR sensor geometry.

\subsection{Overview}

\begin{figure}[t]
    \centering
    \includegraphics[width=0.9\linewidth]{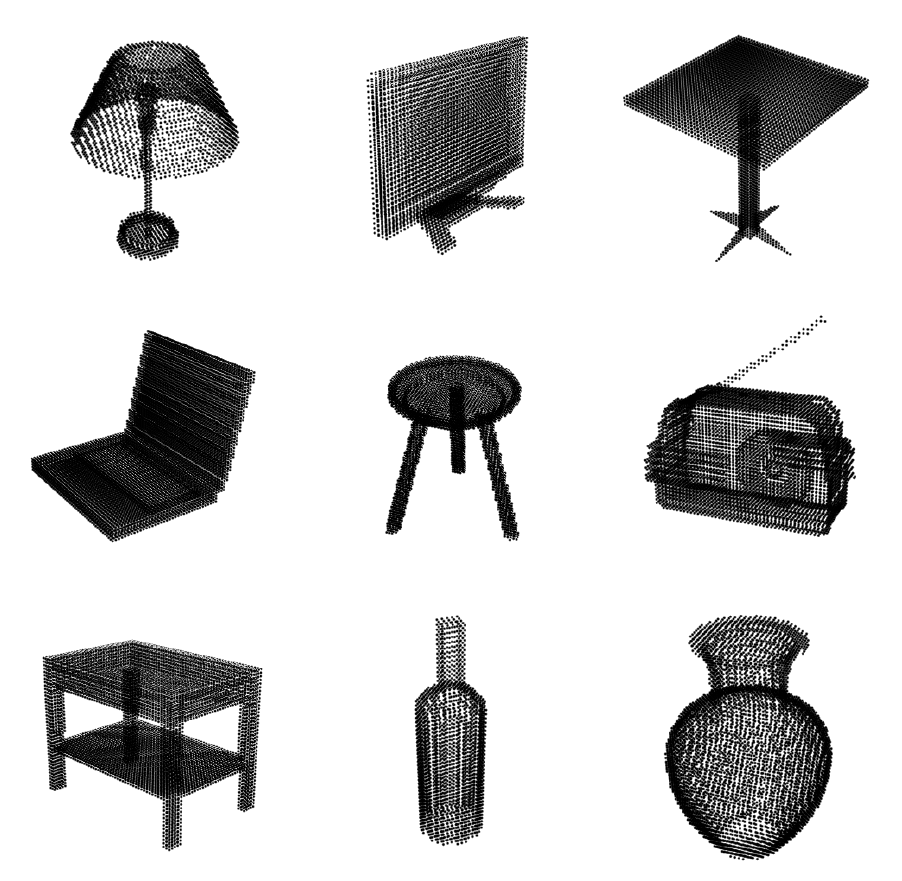}
    \caption{Examples of ModelNet objects selected for the creation of the proposed mixed real-synthetic OoD datasets.}
    \label{fig:modelnet-samples}
\end{figure}

As described in~\cref{sec:datasets}, we construct the proposed Out-of-Distribution (OoD) datasets based on three established autonomous driving benchmarks~\cite{semantickittidataset, pan2020semanticposs, nuscenes}. To insert anomaly objects, we use 3D models from the ModelNet dataset~\cite{wu20153d}, carefully selecting those that do not conflict with the training or evaluation data and are not present in the original training sets, ensuring a totally diverse domain. The selected objects are listed in~\cref{tab:datasets-info}, while others, such as cars, persons, or objects unsuitable for the driving context, like airplanes and guitars, are excluded.
Some objects that are not typically suited for the driving environment, such as bookshelves, dressers, range hoods, or wardrobes, are still included and appropriately resized when inserted to simulate debris, obstacles and other unexpected impediments on the road. This way, we can simulate anomalies that may occur in real-world scenarios, where the vehicle must identify and avoid them. Examples of selected models used for the OoD datasets creation are shown in~\cref{fig:modelnet-samples}.

For each object, we uniformly sample 3D points across the surface of its CAD model to obtain a dense point representation. This facilitates the subsequent alignment of the object's point distribution to the LiDAR scan geometry in the insertion strategy, as described in~\cref{sec:datasets}. Since ModelNet models do not have intensity information, we assign a temporary value of 0 to each point to obtain the same feature configuration of LiDAR scans. We also apply a simple scaling augmentation to randomly reduce the size of the objects, increasing variability.

For each base dataset, we select different surfaces for insertion based on the \textit{single} and \textit{multi} splits introduced in~\cref{sec:datasets}. Specifically, for SemanticKITTI-OoD, anomaly objects in the \textit{single} split are inserted only on the road class label, while the \textit{multi} split also considers classes as parking, sidewalk, and other-ground. For SemanticPOSS-OoD, due to the limited number of class labels, ground is used as the insertion surface in both splits. In nuScenes-OoD, the \textit{single} split contains anomaly objects only on the drivable surface class, whereas the \textit{multi} split also considers other-flat and sidewalk classes.

\begin{figure*}
    \centering
    \includegraphics[width=0.9\linewidth]{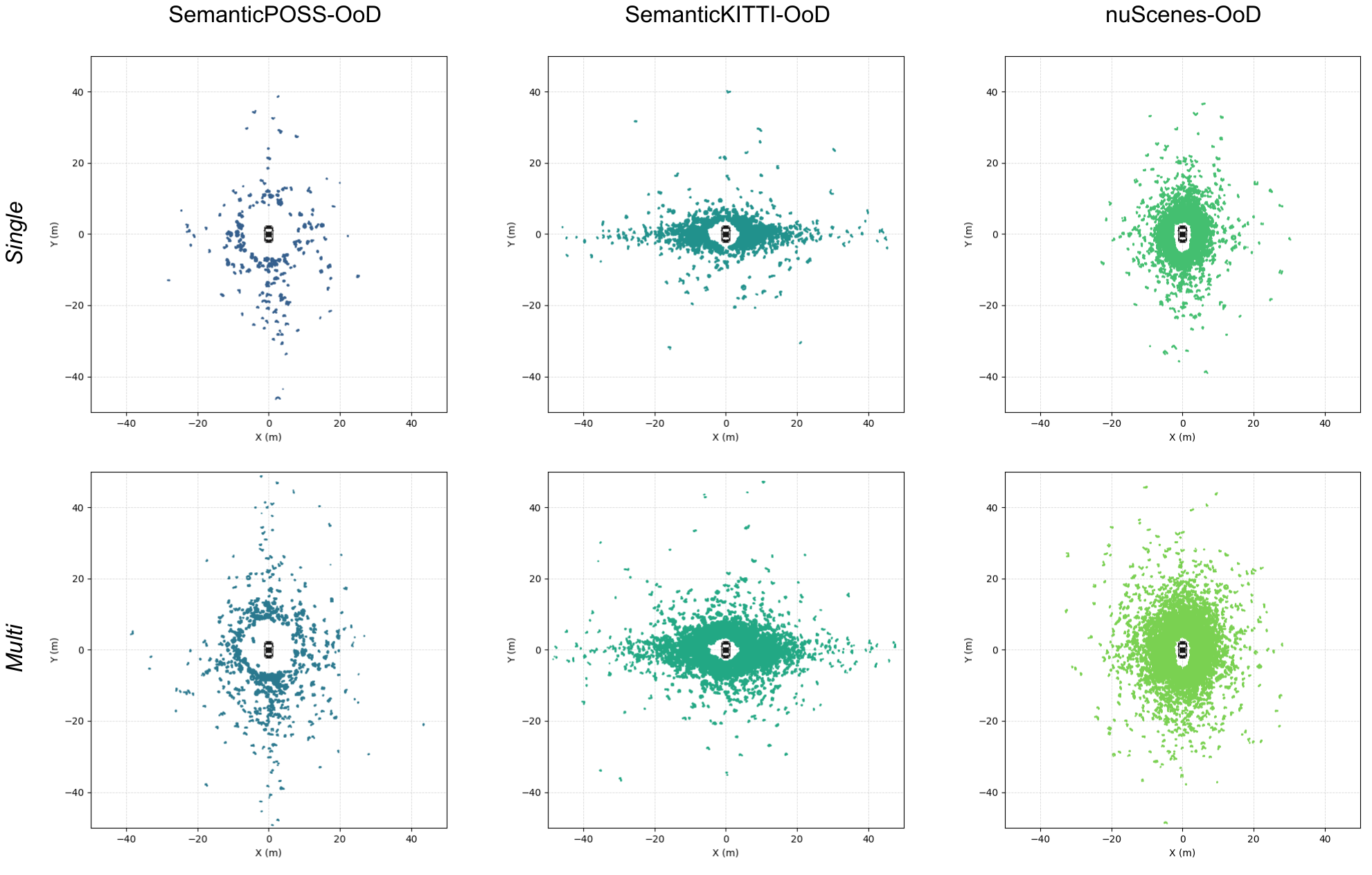}
    \caption{Distribution of anomaly points on the XY plane across all proposed OoD datasets.}
    \label{fig:anomaly-plot}
\end{figure*}

For the \textit{multi} split, we select the number of inserted objects with decreasing probability of 40\%, 30\%, 20\%, and 10\% for inserting 1, 2, 3, and 4 objects, respectively. All anomaly objects are placed within a 50\,m radius from the center of the scan, where the LiDAR sensor is located (see ~\cref{fig:anomaly-plot}), to be consistent with the evaluation setup of~\cite{nekrasov2025spotting}.

Regarding the class labels, we follow~\cite{nekrasov2025spotting} and use the value 2 to denote anomaly points in both SemanticKITTI-OoD and SemanticPOSS-OoD. In nuScenes-OoD, label 2 is already assigned to the \textit{human.pedestrian.adult} class, so we use label 100 for anomaly points to avoid any conflict. 

\subsection{Reflectivity Values}

\begin{table}[t]
    \centering
    \caption{Details of the anomaly objects inserted in the proposed OoD datasets. For each object, we report the number of occurrences in each dataset, along with the total number of anomalies and the per-scan ratio.}
    \label{tab:datasets-info}
    \resizebox{0.47\textwidth}{!}{%
    \begin{tabular}{lcccccc}
        \toprule
        Object & \multicolumn{2}{c}{SemanticKITTI-OoD} & \multicolumn{2}{c}{SemanticPOSS-OoD} & \multicolumn{2}{c}{nuScenes-OoD}\\
         & S & M & S & M & S & M \\
        \midrule
        bathtub      & 59 & 174 & 5 & 17 & 105 & 304 \\
        bed          & 40 & 180 & 5 & 20 & 72 & 239 \\
        bookshelf    & 56 & 158 & 4 & 21 & 77 & 232 \\
        bottle       & 49 & 153 & 5 & 24 & 85 & 273 \\
        bowl         & 66 & 182 & 7 & 15 & 84 & 260 \\
        chair        & 64 & 175 & 9 & 16 & 86 & 251 \\
        cup          & 50 & 168 & 10 & 29 & 77 & 253 \\
        desk         & 60 & 153 & 9 & 22 & 87 & 276 \\
        dresser      & 60 & 177 & 6 & 16 & 76 & 253 \\
        flower pot   & 60 & 161 & 10 & 27 & 73 & 255 \\
        glass box    & 59 & 174 & 2 & 17 & 82 & 232 \\
        lamp         & 55 & 165 & 7 & 18 & 94 & 262 \\
        laptop       & 57 & 171 & 2 & 19 & 75 & 254 \\
        mantel       & 43 & 169 & 9 & 22 & 90 & 223 \\
        monitor      & 60 & 173 & 7 & 24 & 84 & 279 \\
        night stand  & 52 & 175 & 12 & 13 & 87 & 247 \\
        piano        & 60 & 177 & 7 & 21 & 74 & 265 \\
        radio        & 49 & 145 & 6 & 14 & 77 & 263 \\
        range hood   & 57 & 175 & 10 & 22 & 80 & 235 \\
        sink         & 53 & 189 & 6 & 20 & 77 & 212 \\
        sofa         & 67 & 167 & 6 & 20 & 94 & 275 \\
        stool        & 48 & 168 & 11 & 20 & 98 & 258 \\
        table        & 71 & 146 & 9 & 21 & 77 & 249 \\
        tent         & 68 & 164 & 2 & 25 & 88 & 241 \\
        toilet       & 62 & 169 & 8 & 28 & 75 & 232 \\
        tv stand     & 51 & 172 & 5 & 26 & 82 & 242 \\
        vase         & 45 & 165 & 7 & 15 & 85 & 237 \\
        wardrobe     & 61 & 172 & 8 & 19 & 78 & 238 \\
        xbox         & 53 & 177 & 2 & 15 & 79 & 228 \\
        \midrule
        Total & 1634 & 4894 & 196 & 586 & 2398 & 7268 \\
        Ratio & 0.40 & 1.20 & 0.39 & 1.17 & 0.40 & 1.21 \\
        \bottomrule
    \end{tabular}}
\end{table}

\begin{table}[t]
    \centering
    \caption{Material properties and reflectivity values for the selected ModelNet objects.}
    \label{tab:reflectivity}
    \begin{tabular}{l l c}
        \toprule
        Object & Material & Reflectivity \\
        \midrule
        bathtub        & glossy ceramic              & 0.60 \\
        bed            & fabric, wood               & 0.40 \\
        bookshelf      & wood                        & 0.40 \\
        bottle         & glass, plastic               & 0.25 \\
        bowl           & ceramic                     & 0.60 \\
        chair          & wood, plastic                & 0.35 \\
        cup            & ceramic                     & 0.60 \\
        desk           & wood                         & 0.40 \\
        dresser        & wood                        & 0.40 \\
        flower pot     & clay                        & 0.45 \\
        glass box      & glass                       & 0.20 \\
        lamp           & metal, plastic               & 0.35 \\
        laptop         & metal, plastic               & 0.30 \\
        mantel         & stone                       & 0.40 \\
        monitor        & plastic, glass              & 0.25 \\
        night stand    & wood                        & 0.40 \\
        piano          & polished wood               & 0.50 \\
        radio          & plastic, metal               & 0.35 \\
        range hood     & steel                       & 0.45 \\
        sink           & ceramic, metal               & 0.60 \\
        sofa           & fabric                      & 0.40 \\
        stool          & wood, metal                  & 0.35 \\
        table          & wood, plastic               & 0.40 \\
        tent           & fabric                      & 0.30 \\
        toilet         & glossy ceramic              & 0.60 \\
        tv stand       & wood, plastic                & 0.40 \\
        vase           & ceramic                     & 0.55 \\
        wardrobe       & wood                        & 0.40 \\
        xbox           & plastic                     & 0.30 \\
        \bottomrule
    \end{tabular}
\end{table}

As introduced in~\cref{sec:datasets}, ModelNet objects do not provide intensity information, while real-world LiDAR scans usually have this attribute. To solve the mismatch and ensure consistency between inserted objects and the scene, we compute a per-point intensity value (see~\cref{eq:intesity-formula}) that approximates real-world behavior following the Lambertian reflectance model~\cite{oren1995generalization}. This requires assigning a reflectivity value $\rho$ to each object, representing the intrinsic reflectance of its material.

\begin{figure}[t]
    \centering
    \includegraphics[width=0.99\linewidth]{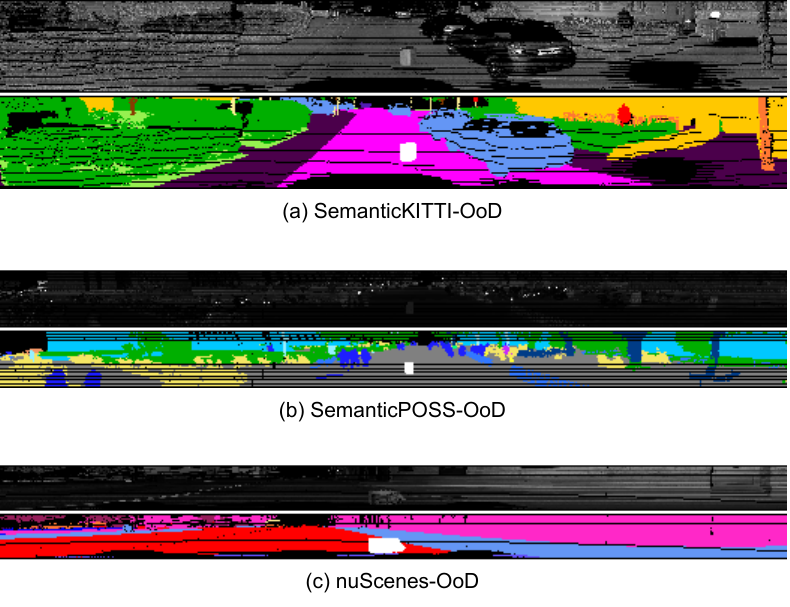}
    \caption{Example of computed intensity values in the proposed OoD datasets. (Top) Remission value of the LiDAR scan. (Bottom) Semantic labels, with inserted anomaly objects shown in white.}
    \label{fig:intensity-sample}
\end{figure}

We assign material types to the selected ModelNet objects based on plausible real-world composition, for example, treating chairs as wood or plastic and vases as ceramic. We assign reflectivity values accordingly, based on empirical observations and conventions used in rendering engines (e.g., Blender).
Materials such as glass, ice or dark surfaces show low reflectance, while mirrors or glossy metals reflect significantly more light, resulting in higher reflectivity values. These values range in $[0, 1]$ where 0 and 1 indicate respectively the minimum and maximum reflectance. \Cref{tab:reflectivity} reports assigned materials and corresponding reflectivity values for all selected models. ~\Cref{fig:intensity-sample} shows some examples of the computed intensity values obtained using the proposed technique, visualized on sections of the range image projections. In SemanticKITTI-OoD, the effect of the intensity computation on the inserted object is more evident, while in the other two datasets, the low mean intensity value makes anomalies more difficult to distinguish from the background, particularly for SemanticPOSS-OoD. Overall, results demonstrate that the proposed approach for intensity estimation adapts to the characteristics of corresponding LiDAR scans, producing values that correctly align with the sensor behavior.

\subsection{Insights on Protocol}

We provide additional details on the proposed strategy for creating the OoD datasets, starting from each base autonomous driving benchmark. Given the combined point cloud $\mathbf{S} = [\mathbf{P}, \mathbf{O}]$, where $\mathbf{P} \in \mathbb{R}^{N \times 4}$ is the LiDAR scan containing $N$ points and $\mathbf{O} \in \mathbb{R}^{M \times 4}$ the object point cloud with $M$ points, we perform a spherical projection of 3D points onto a 2D surface, namely, a range image~\cite{milioto2019rangenet++}. This projection provides essential geometric information for preserving realistic scan properties, such  as point occlusion and the beam-like point distribution characteristic of LiDAR measurements.
Each point $\mathbf{p}_n \in \mathbf{S}$ where $\mathbf{p}_n = (x_n, y_n, z_n, i_n)$ with coordinates and intensity value, is projected onto a range image $R(u,v) \in \mathbb{R}^{H \times W}$, where $H$ and $W$ are the height and width. We set $H$ as the number of beams of the LiDAR sensor, for each dataset, and $W = 2048$ for a wider horizontal resolution, to keep more anomaly points. The projection procedure is as follows:

\begin{equation}\label{eq:spherical-projection}
	\binom{u_n}{v_n} = \binom{\frac{1}{2}[1 - \arctan(y_n, x_n)\pi^{-1}]W}{[1 - (\arcsin(z_n, r_n^{-1}) + f_\text{down})f^{-1}]H},
\end{equation}

where $r_n = \sqrt{x_n^2 + y_n^2 + z_n^2}$ is the range of the point with respect to the sensor, $f = |f_\text{up} + f_\text{down}|$ denotes the vertical field of view of the sensor, and $f_\text{up}$, $f_\text{down}$ are the upward and downward inclination angles, respectively. In the case of multiple points projected onto the
same cell, we select the closest point to the sensor, i.e., the one having the minimum range value.

Each cell of the range image stores the range value of the 3D point projected in the corresponding cell, providing essential information for dataset construction and for geometrically aligning the point distribution of the inserted objects with that of the LiDAR scans. 
First, it provides a direct mapping between 3D points and 2D cells, allowing us to identify points in the original LiDAR scan $\mathbf{S}$ that become occluded by the inserted anomaly object, and should be removed in the final scan.
Second, for points belonging to the inserted objects, rows of the range image correspond to LiDAR beams. We select points along each row, obtaining a beam-like sampling pattern for the object, aligned with the LiDAR data acquisition process. Furthermore, since each cell retains only the closest point to the sensor, this process ensures that only visible, front points of the object are preserved. The projection and reprojection operations also avoid any possible overlap of the inserted objects with instances in the LiDAR scan, maintaining a geometrically consistent real-world scenario.

\section{Further Details on Experiments}

\subsection{Baselines}

Due to the novelty of the task, as discussed in~\cref{sec:experiments}, there are only a few methods developed for 3D LiDAR anomaly segmentation. Moreover, only a subset of these approaches provides publicly available code~\cite{nekrasov2025spotting}. Therefore, for the experiments on the proposed OoD datasets, we rely on the implementation from~\cite{nekrasov2025spotting}, which, however, includes only a portion of the methods tested in that work, namely max logit, RbA, and deep ensemble.
For a fair comparison, we additionally evaluate standard OoD methods considered in~\cite{nekrasov2025spotting}, using the same backbone as our approach, i.e., MinkowskiNet~\cite{choy20194d}, to assess any possible bias related to the backbone. We retrain these methods on the corresponding training split of each OoD dataset and evaluate them using the same inference setting.

For Void Classifier, we train the network with an additional class corresponding to the unlabeled/outlier regions and use its predicted confidence at inference time to identify anomalies. For MC Dropout, we introduce dropout layers in the backbone during training and activate them in inference, where multiple forward pass are repeated (10 in our setup). For the deep ensemble, following~\cite{nekrasov2025spotting}, we train three models with different random seeds to obtain different checkpoints. We then extract the pre-softmax per-point features from each model and combine them by computing the mean and then calculating the entropy of the prediction probabilities to produce the final anomaly segmentation scores.

\begin{table}[t]
    \centering
    \caption{Ablation study on the threshold $r$ in the contrastive head score~\cref{eq:score-contrastive}. Results on STU validation set.}
    \label{tab:ablation-threshold}
    \resizebox{0.47\textwidth}{!}{%
    \begin{tabular}{cccc}
        \toprule
        Threshold $r$ & AUROC [\%] $\uparrow$ & FPR@95 [\%] $\downarrow$ & AP [\%] $\uparrow$ \\
        \midrule
        1 & 93.17 & 36.47 & 18.87 \\
        2 & 94.51 & 34.66 & 26.92 \\
        3 & 94.79 & \textbf{34.43} & 27.04 \\
        4 & 94.90 & 34.65 & 27.46 \\
        \rowcolor{green!15}
        5 & \textbf{95.05} & 34.86 & \textbf{27.53} \\
        \bottomrule
    \end{tabular}}
\end{table}

\begin{table}[t]
    \centering
    \caption{Ablation study of the LiDAR anomaly segmentation approach pipeline on SemanticKITTI-OoD (\textit{Multi}) set, analyzing its impact on semantic segmentation.}
    \label{tab:ablation-suppl}
    \begin{tabular}{cccc}
        \toprule
        $\mathcal{L}_\text{prot}$ & $\mathcal{L}_\text{cont}$ & $\mathcal{L}_\text{obj}$ & mIoU (\%) \\
        \midrule
         &  &  & \textbf{64.8} \\
        \checkmark &  &  & 60.7 \\
        \checkmark & \checkmark &  & 61.0 \\
        \checkmark & \checkmark & \checkmark & 61.2 \\
        \bottomrule
    \end{tabular}
\end{table}

\begin{table}[t]
    \centering
    \caption{Model runtime comparison across different datasets. Results are in reported in\,ms. Tested on a NVIDIA A40 GPU.}
    \label{tab:suppl-runtime}
    \resizebox{0.47\textwidth}{!}{%
    \begin{tabular}{lcccc}
        \toprule
        Method & KITTI-OoD & POSS-OoD & nuScenes-OoD & STU \\
        \midrule
        Mask4Former3D~\cite{yilmaz2024mask4former} & 348 & 315 & 168 & 392 \\
        Deep Ensemble (Sequential)~\cite{lakshminarayanan2017simple} & 2454 & 1695 & 861 & 2628 \\
        Deep Ensemble (Parallel)~\cite{lakshminarayanan2017simple} & 818 & 565 & 287 & 876 \\
        \rowcolor{green!15}
        \textbf{LIDO (ours)} & \textbf{87} & \textbf{75} & \textbf{38} & \textbf{90} \\
        \bottomrule
    \end{tabular}}
\end{table}

\begin{table}[t]
    \centering
    \caption{Range-based evaluation on STU validation set (AP, \%).}
    \label{tab:range-evaluation}
    \resizebox{0.46\textwidth}{!}{%
    \begin{tabular}{lccccc}
        \toprule
        Method & 0-10m & 10-20m & 20-30m & 30-40m & 40-50m \\
        \midrule
        RbA~\cite{nayal2023rba} & 1.85 & 1.28 & 0.73 & 0.15 & 0.01 \\
        Max Logit~\cite{hendrycks2017baseline} & 2.25 & 1.53 & 1.20 & 0.27 & 0.01 \\
        Deep Ensemble~\cite{lakshminarayanan2017simple} & 7.63 & 8.49 & 3.42 & \textbf{0.38} & \textbf{0.03} \\
        MC Dropout~\cite{srivastava2014dropout} & 0.16 & 0.53 & 0.06 & 0.04 & 0.01 \\
        Void Classifier~\cite{blum2021fishyscapes} & 2.95 & 1.78 & 1.98 & 0.28 & \textbf{0.03} \\
        \rowcolor{green!15}
        \textbf{LIDO (ours)} & \textbf{40.77} & \textbf{44.07} & \textbf{5.51} & 0.21 & 0.02 \\
        \bottomrule
    \end{tabular}}
\end{table}

\subsection{Additional Ablation Study}

In~\cref{tab:ablation-threshold}, we report results obtained with different threshold values $r$, as defined in~\cref{eq:score-contrastive}. Due to the large domain gap between training and evaluation, when training solely on SemanticKITTI and testing on the STU dataset, this parameter requires tuning. Differences in sensor resolution and feature distributions may produce variations in the feature norms. Unlike findings in the image domain~\cite{sodano2024cvpr}, we find that feature norms of inlier classes, produced from LiDAR scans, tends to be larger. This explains the choice of a higher threshold, as reported in the table, with $r=5$ obtaining the best performance. Other values have slightly lower FPR, but do not match the best AUROC and AP metrics.

We also analyze the impact of the loss functions introduced in our approach on semantic segmentation performance (mIoU). Results are reported in~\cref{tab:ablation-suppl} on the proposed SemanticKITTI-OoD dataset, \textit{multi} split, as STU~\cite{nekrasov2025spotting} validation set does not provide semantic labels. We observe that introducing the prototype loss $\mathcal{L}_\text{prot}$ leads to a  decrease in mIoU, while the addition of contrastive and objectosphere losses slightly recovers this drop, by improving the separation between class-specific features. Although this results in a slight reduction in semantic segmentation results, it yields improved anomaly segmentation results, consistent with the trends observed in the ablation study on STU (see~\cref{sec:experiments}).


\subsection{Runtime} 

We report a runtime comparison between our proposed approach and other baselines~\cite{nekrasov2025spotting} in~\cref{tab:suppl-runtime}, on the STU dataset and the proposed real-synthetic OoD datasets. All methods are tested on a single NVIDIA A40 GPU. The results show that our approach maintains real-time performance ($<$100\,ms) across all datasets, including high-resolution configurations such as the 128-beam STU scans. Instead, Mask4Former3D~\cite{nekrasov2025spotting} combined with post-processing techniques as max logit or RbA, and especially when used in an ensemble setting, drastically increases the runtime, requiring seconds to produce a single prediction.

\subsection{Additional Results}

We compute the AP metric across different distance thresholds and report results in~\cref{tab:range-evaluation}, following the protocol of~\cite{nekrasov2025spotting}. We compare our approach with the Mask4Former3D baselines introduced in~\cite{nekrasov2025spotting}. As expected, our method achieves superior performance at shorter ranges, with a decrease as the distance to anomalous objects increases, a trend observed across all approaches. These results highlight the effectiveness of our approach in identifying anomalies while still indicating room for improvements for distant objects, which remain a challenging setting.

We report extensive results on both STU and the proposed mixed real-synthetic OoD m, datasets in~\cref{tab:stu-suppl,tab:poss-suppl,tab:kitti-suppl,tab:nuscenes-suppl}, including additional baselines evaluated with the same backbone as our method, i.e., MinkowskiNet~\cite{choy20194d}, for a fair and more comprehensive comparison. \Cref{tab:stu-suppl} presents results on the STU validation set, including object-level OoD metrics from~\cite{nekrasov2025spotting} for a better evaluation of anomaly segmentation performance, such as Panoptic Quality (PQ), Unknown Quality (UQ)~\cite{wong2020identifying}, Recognition Quality (RQ) and Segmentation Quality (SQ). We refer the reader to the original paper for more details on these metrics. Our method significantly outperforms all baselines in terms of AP, indicating strong capability in identifying anomalous objects, and also achieves overall better object-level performance. In contrast, applying standard OoD technique on top of the same MinkowskiNet backbone yield comparable or even inferior results compared to the Mask4Former3D baseline from~\cite{nekrasov2025spotting}.
Results on the other datasets further confirm the effectiveness of the proposed approach, consistently achieving strong performance. While standard OoD methods based on either Mask4Former3D or MinkowskiNet remain competitive, they generally do not surpass our method. When they do so, as in the case of Deep Ensemble, they incur in significantly higher computational and memory costs.
nuScenes-OoD (\cref{tab:nuscenes-suppl}) represents a particular challenging scenario, mainly due to the lower number of LiDAR beams and reduced point density. As discussed in~\cref{sec:experiments}, these factors affect the learning of good and robust class prototypes, which also reflects in the observed drop in semantic segmentation performance. In this dataset, our method achieves comparable results with standard OoD methods with the same backbone in terms of AP, while maintaining lower FPR values, highlighting the effectiveness despite the challenging setting and indicating room for further improvements in generalization. 


\begin{table*}[t]
    \caption{Anomaly Segmentation performance on STU validation set.}
    \label{tab:stu-suppl}
    \centering
    \resizebox{0.99\linewidth}{!}{
    \begin{tabular}{lccccccccc}
        \toprule
         & \multicolumn{3}{c}{Point-Level OoD} & \multicolumn{5}{c}{Object-Level OoD} \\
        Method & AUROC [\%] $\uparrow$ & FPR@95 [\%] $\downarrow$ & AP [\%] $\uparrow$ & RecallQ [\%] & SQ [\%] & RQ [\%] & UQ [\%] & PQ [\%] \\
         \midrule
        Mask4Former3D + MC Dropout~\cite{srivastava2014dropout} & 65.76 & 79.82 & 0.17 & 3.54 & 74.36 & 3.48 & 2.63 & 2.59 \\
        Mask4Former3D + RbA~\cite{nayal2023rba} & 73.00 & 100.0 & 1.64 & 21.84 & 78.58 & 2.75 & 17.16 & 2.16 \\
        Mask4Former3D + Max Logit~\cite{hendrycks2017baseline} & 87.27 & 68.76 & 2.02 & 26.64 & 79.26 & 2.06 & 21.12 & 1.63 \\
        Mask4Former3D + Void Classifier~\cite{blum2021fishyscapes} & 89.77 & 79.50 & 2.62 & 17.35 & \underline{81.27} & 8.98 & 14.10 & \underline{7.30} \\
        Mask4Former3D + Deep Ensemble~\cite{lakshminarayanan2017simple} & 90.93 & 37.34 & \underline{6.94} & 17.70 & 79.96 & \underline{9.10} & 14.15 & 7.27 \\
        MinkowskiNet + RbA~\cite{nayal2023rba} & 53.93 & 92.07 & 0.05 & 4.76 & \textbf{88.49} & 0.26 & 4.21 & 0.23 \\
        MinkowskiNet + Void Classifier~\cite{blum2021fishyscapes} & 70.24 & 100.0 & 0.11 & - & - & - & - & - \\
        MinkowskiNet + Max Logit~\cite{hendrycks2017baseline} & 90.92 & 43.49 & 0.97 & - & - & - & - & - \\
        MinkowskiNet + MC Dropout~\cite{srivastava2014dropout} & 91.98 & 35.64 & 1.41 & \underline{35.25} & 75.83 & 1.35 & \underline{26.73} & 1.02 \\
        MinkowskiNet +  Deep Ensemble~\cite{lakshminarayanan2017simple} & \underline{93.32} & \textbf{29.67} & 2.72 & \textbf{39.88} & 78.40 & 2.01 & \textbf{31.27} & 1.57 \\
        \rowcolor{green!15}
        \textbf{LIDO (ours)} & \textbf{95.05} & \underline{34.86} & \textbf{27.53} & 31.92 & 74.82 & \textbf{28.38} & 23.88 & \textbf{21.23} \\
         \bottomrule
    \end{tabular}}
\end{table*}

\begin{table*}[t]
    \caption{Anomaly Segmentation performance on SemanticPOSS-OoD dataset.}
    \label{tab:poss-suppl}
    \centering
    \resizebox{0.99\linewidth}{!}{
    \begin{tabular}{lcccccc}
        \toprule
         & \multicolumn{3}{c}{Single} & \multicolumn{3}{c}{Multi} \\
        Method & AUROC [\%] $\uparrow$ & FPR@95 [\%] $\downarrow$ & AP [\%] $\uparrow$ & AUROC [\%] $\uparrow$ & FPR@95 [\%] $\downarrow$ & AP [\%] $\uparrow$ \\
         \midrule
        Mask4Former3D + RbA~\cite{nayal2023rba} & 49.09 & 100.0 & 0.13 & 50.42 & 100.0 & 0.26 \\
        Mask4Former3D + Max Logit~\cite{hendrycks2017baseline} & 61.05 & 91.85 & 0.19 & 63.77 & 88.74 & 0.40 \\
        Mask4Former3D + Deep Ensemble~\cite{lakshminarayanan2017simple} & 85.86 & 58.12 & 0.82 & 85.83 & 54.20 & 1.21 \\
        MinkowskiNet + RbA~\cite{nayal2023rba} & 22.21 & 99.32 & 0.08 & 19.97 & 99.43 & 0.14 \\
        MinkowskiNet + Void Classifier~\cite{blum2021fishyscapes} & 46.22 & 98.38 & 0.12 & 49.02 & 97.30 & 0.24 \\
        MinkowskiNet + Max Logit~\cite{hendrycks2017baseline} & 87.15 & 48.21 & 1.36 & 89.86 & 47.72 & 2.94 \\
        MinkowskiNet + MC Dropout~\cite{srivastava2014dropout} & 85.49 & 66.30 & 1.41 & 86.11 & 58.76 & 2.93 \\
        MinkowskiNet + Deep Ensemble~\cite{lakshminarayanan2017simple} & \underline{89.82} & \underline{45.91} & \underline{2.54} & \textbf{91.15} & \underline{46.33} & \underline{4.87} \\
        \rowcolor{green!15}
        \textbf{LIDO (ours)} & \textbf{91.51} & \textbf{45.10} & \textbf{3.97} & \underline{90.84} & \textbf{44.74} & \textbf{5.92} \\
         \bottomrule
    \end{tabular}}
\end{table*}

\begin{table*}[t]
    \caption{Anomaly Segmentation performance on SemanticKITTI-OoD dataset.}
    \label{tab:kitti-suppl}
    \centering
    \resizebox{0.99\linewidth}{!}{
    \begin{tabular}{lcccccc}
        \toprule
         & \multicolumn{3}{c}{Single} & \multicolumn{3}{c}{Multi} \\
        Method & AUROC [\%] $\uparrow$ & FPR@95 [\%] $\downarrow$ & AP [\%] $\uparrow$ & AUROC [\%] $\uparrow$ & FPR@95 [\%] $\downarrow$ & AP [\%] $\uparrow$ \\
         \midrule
        Mask4Former3D + RbA~\cite{nayal2023rba} & 59.68 & 100.0 & 4.56 & 71.39 & 100.0 & \underline{13.63} \\
        Mask4Former3D + Max Logit~\cite{hendrycks2017baseline} & 76.49 & 99.63 & 4.84 & 84.05 & 99.32 & \textbf{13.84} \\
        Mask4Former3D + Deep Ensemble~\cite{lakshminarayanan2017simple} & \underline{92.87} & \textbf{27.69} & \underline{6.20} & \textbf{92.19} & \textbf{28.64} & 12.04 \\
        MinkowskiNet + RbA~\cite{nayal2023rba} & 40.38 & 95.45 & 0.19 & 36.90 & 96.54 & 0.41 \\
        MinkowskiNet + Void Classifier~\cite{blum2021fishyscapes} & 47.74 & 87.02 & 0.22 & 50.89 & 84.33 & 0.53 \\
        MinkowskiNet + Max Logit~\cite{hendrycks2017baseline} & 88.70 & 52.26 & 2.49 & 86.93 & 48.01 & 3.88 \\
        MinkowskiNet + MC Dropout~\cite{srivastava2014dropout} & 87.49 & 50.75 & 3.04 & 84.93 & 57.46 & 4.48 \\
        MinkowskiNet + Deep Ensemble~\cite{lakshminarayanan2017simple} & 91.64 & 39.96 & 4.78 & 89.03 & 42.34 & 6.21 \\
        \rowcolor{green!15}
        \textbf{LIDO (ours)} & \textbf{93.36} & \underline{31.19} & \textbf{10.60} & \underline{89.89} & \underline{39.04} & 9.42 \\
         \bottomrule
    \end{tabular}}
\end{table*}

\begin{table*}[t]
    \caption{Anomaly Segmentation performance on nuScenes-OoD dataset.}
    \label{tab:nuscenes-suppl}
    \centering
    \resizebox{0.99\linewidth}{!}{
    \begin{tabular}{lcccccc}
        \toprule
         & \multicolumn{3}{c}{Single} & \multicolumn{3}{c}{Multi} \\
        Method & AUROC [\%] $\uparrow$ & FPR@95 [\%] $\downarrow$ & AP [\%] $\uparrow$ & AUROC [\%] $\uparrow$ & FPR@95 [\%] $\downarrow$ & AP [\%] $\uparrow$ \\
         \midrule
        Mask4Former3D + RbA~\cite{nayal2023rba} & 32.65 & 100.0 & 3.56 & 71.60 & 98.68 & 7.95 \\
        Mask4Former3D + Max Logit~\cite{hendrycks2017baseline} & 66.17 & 98.60 & 3.54 & 43.54 & 100.0 & 8.98 \\
        Mask4Former3D + Deep Ensemble~\cite{lakshminarayanan2017simple} & \textbf{91.79} & \textbf{34.59} & \textbf{18.34} & \underline{88.62} & \underline{43.63} & \textbf{23.87} \\
        MinkowskiNet + RbA~\cite{nayal2023rba} & 11.02 & 99.59 & 0.41 & 12.61 & 99.52 & 0.80 \\
        MinkowskiNet + MC Dropout~\cite{srivastava2014dropout} & 84.14 & 50.80 & 4.48 & 81.94 & 53.09 & 6.64 \\
        MinkowskiNet + Max Logit~\cite{hendrycks2017baseline} & \underline{90.82} & 41.11 & 8.14 & \textbf{90.10} & \textbf{38.73} & 12.11 \\
        MinkowskiNet + Deep Ensemble~\cite{lakshminarayanan2017simple} & 89.79 & \underline{39.28} & 8.26 & 87.29 & 44.37 & 11.30 \\
        MinkowskiNet + Void Classifier~\cite{blum2021fishyscapes} & 86.09 & 61.40 & \underline{8.30} & 86.64 & 61.57 & \underline{16.34} \\
        \rowcolor{green!15}
        \textbf{LIDO (ours)} & 89.33 & 39.70 & 6.79 & 87.25 & 44.51 & 10.53 \\
         \bottomrule
    \end{tabular}}
\end{table*}


\begin{table*}[t]
    \centering
    \caption{LiDAR Semantic Segmentation results on STU inlier validation sequence.}
    \label{tab:semseg-stu}
    \resizebox{0.99\textwidth}{!}{%
    \begin{tabular}{l>{\columncolor{green!15}}cccccccccccccccc}
        \toprule
        Method & \rb{mIoU\,\%} & \rb{car} & \rb{bicycle} & \rb{truck} & \rb{person} & \rb{road} & \rb{parking} & \rb{sidewalk} & \rb{building} & \rb{fence} & \rb{vegetation}  & \rb{trunk} & \rb{terrain} & \rb{pole} & \rb{traffic-sign} \\ 
        \midrule
        Standard & \textbf{36.7} & 62.4 & 0.0 & 2.4 & 49.8 & 67.3 & 16.0 & 57.0 & \textbf{84.1} & \textbf{75.2} & \textbf{90.9} & \textbf{34.7} & 67.1 & \textbf{45.7} & \textbf{45.8} \\
        \textbf{LIDO (ours)} & 35.1 & \textbf{70.3} & \textbf{0.8} & \textbf{53.1} & \textbf{67.6} & \textbf{73.2} & \textbf{19.4} & \textbf{59.7} & 73.7 & 70.1 & 66.8 & 21.4 & \textbf{67.7} & 12.1 & 11.9 \\
        \bottomrule
    \end{tabular}}
\end{table*}

\begin{table*}[t]
    \centering
    \caption{LiDAR Semantic Segmentation results on the proposed SemanticKITTI-OoD dataset.}
    \label{tab:semseg-kitti}
    \resizebox{0.99\textwidth}{!}{%
    \begin{tabular}{ll>{\columncolor{green!15}}cccccccccccccccccccc}
        \toprule
         & Method & \rb{mIoU\,\%} & \rb{car} & \rb{bicycle} & \rb{motorcycle} & \rb{truck} & \rb{other-vehicle} &  \rb{person} & \rb{bicyclist} & \rb{motorcyclist} & \rb{road} & \rb{parking} & \rb{sidewalk} & \rb{other-ground} & \rb{building} & \rb{fence} & \rb{vegetation}  & \rb{trunk} & \rb{terrain} & \rb{pole} & \rb{traffic-sign} \\ 
        \midrule
        \multirow{2}{*}{\rb{\scriptsize Single}} & Standard & \textbf{64.9} & \textbf{96.8} & \textbf{32.6} & \textbf{80.8 }& \textbf{78.6} & \textbf{61.6} & \textbf{71.6} & \textbf{89.8} & \textbf{0.1} & \textbf{93.5} & \textbf{52.0} & \textbf{81.0} & 0.2 & \textbf{91.0} & \textbf{60.2} & \textbf{87.9} & \textbf{67.0} & \textbf{75.7} & \textbf{64.9} & 49.7 \\
         & \textbf{LIDO (ours)} & 61.3 & 95.9 & 18.7 & 74.2 & 74.5 & 61.0 & 71.1 & 82.8 & 0.0 & 91.4 & 42.4 & 77.0 & \textbf{0.8} & 89.5 & 52.6 & 86.4 & 66.3 & 74.0 & 55.7 & \textbf{51.0} \\
        \midrule
        \multirow{2}{*}{\rb{\scriptsize Multi}}
         & Standard & \textbf{64.8} & \textbf{96.9} & \textbf{32.7} & \textbf{80.9} & \textbf{77.8} & \textbf{60.8} & \textbf{71.4} & \textbf{89.7} & \textbf{0.1} & \textbf{93.4} & \textbf{51.5} & \textbf{80.8} & 0.2 & \textbf{90.9} & \textbf{59.9} & \textbf{87.9} & \textbf{67.0} & \textbf{75.7} & \textbf{64.8} & 49.7 \\
         & \textbf{LIDO (ours)} & 61.2 & 95.9 & 18.7 & 73.3 & 74.4 & \textbf{60.8} & 70.8 & 82.7 & 0.0 & 91.2 & 42.2 & 76.6 & \textbf{0.8} & 89.5 & 52.4 & 86.4 & 66.3 & 74.0 & 55.7 & \textbf{50.9} \\
        \bottomrule
    \end{tabular}}
\end{table*}

\begin{table*}[t]
    \centering
    \caption{LiDAR Semantic Segmentation results on the proposed SemanticPOSS-OoD dataset.}
    \label{tab:semseg-poss}
    \resizebox{0.99\textwidth}{!}{%
    \begin{tabular}{ll>{\columncolor{green!15}}cccccccccccccc}
        \toprule
          & Method & \rb{mIoU\,\%} & \rb{person} & \rb{rider} & \rb{car} & \rb{truck} & \rb{plants} & \rb{traffic sign} & \rb{pole} & \rb{trashcan} & \rb{building} & \rb{cone/stone} & \rb{fence} & \rb{bike} & \rb{ground} \\
        \midrule
        \multirow{2}{*}{\rb{\scriptsize Single}} & Standard & \textbf{57.1} & 78.0 & \textbf{34.6} & 80.7 & \textbf{33.5} & \textbf{74.7} & 24.4 & \textbf{41.1} & \textbf{65.4} & 79.5 & 40.9 & \textbf{51.0} & \textbf{57.9} & \textbf{80.2} \\
         & \textbf{LIDO (ours)} & 55.6 & \textbf{79.1} & 32.2 & \textbf{82.9} & 21.6 & 67.9 & \textbf{27.1} & 34.4 & 58.9 & \textbf{81.5} & \textbf{57.6} & 49.0 & 53.6 & 77.3 \\
        \midrule
        \multirow{2}{*}{\rb{\scriptsize Multi}}
         & Standard & \textbf{57.0} & 78.0 & \textbf{34.5} & 80.7 & \textbf{33.5} & \textbf{74.7} & 24.4 & \textbf{41.1} & \textbf{65.3} & 79.5 & 40.9 & \textbf{51.0} & \textbf{57.9} & \textbf{80.2} \\
         & \textbf{LIDO (ours)} & 55.5 & \textbf{79.1} & 32.2 & \textbf{82.9} & 21.6 & 67.9 & \textbf{27.1} & 34.4 & 58.4 & \textbf{81.5} & \textbf{57.5} & 49.0 & 53.6 & 77.2 \\
        \bottomrule
    \end{tabular}}
\end{table*}

\begin{table*}[t]
    \centering
    \caption{LiDAR Semantic Segmentation results on the proposed nuScenes-OoD dataset.}
    \label{tab:semseg-nuscenes}
    \resizebox{0.99\textwidth}{!}{%
    \begin{tabular}{ll>{\columncolor{green!15}}ccccccccccccccccc}
        \toprule
          & Method & \rb{mIoU\,\%} & \rb{barrier} & \rb{bicycle} & \rb{bus} & \rb{car} & \rb{const. veh.} & \rb{motorcycle} & \rb{pedestrian} & \rb{traffic cone} & \rb{trailer} & \rb{truck} & \rb{driv. surf.} & \rb{other flat} & \rb{sidewalk} & \rb{terrain} & \rb{manmade} & \rb{vegetation} \\
        \midrule
        \multirow{2}{*}{\rb{\scriptsize Single}} & Standard & \textbf{72.7} & \textbf{74.3} & \textbf{40.2} & \textbf{87.9} & \textbf{90.5} & \textbf{40.9} & \textbf{81.8} & \textbf{76.8} & \textbf{60.0} & \textbf{54.2} & \textbf{81.1} & \textbf{95.3} & 66.1 & \textbf{70.2} & \textbf{74.1} & \textbf{85.9} & \textbf{84.7} \\
         & \textbf{LIDO (ours)} & 60.6 & 57.7 & 5.1 & 70.8 & 88.0 & 16.2 & 51.1 & 76.0 & 26.4 & 36.5 & 77.6 & 94.4 & \textbf{66.4} & 67.6 & 72.1 & 81.7 & 82.0 \\
        \midrule
        \multirow{2}{*}{\rb{\scriptsize Multi}}
         & Standard & \textbf{72.6} & \textbf{90.5} & \textbf{40.1} & \textbf{87.9} & \textbf{90.5} & \textbf{40.7} & \textbf{81.7} & \textbf{76.8} & \textbf{59.2} & \textbf{54.2} & \textbf{81.1} & \textbf{95.2} & 65.6 & \textbf{69.9} & \textbf{74.0} & \textbf{85.9} & \textbf{84.7} \\
         & \textbf{LIDO (ours)} & 60.4 & 57.6 & 5.1 & 70.8 & 88.1 & 16.2 & 51.2 & 76.1 & 25.4 & 36.6 & 77.6 & 94.1 & \textbf{65.7} & 66.9 & 72.0 & 81.7 & 82.0 \\
        \bottomrule
    \end{tabular}}
\end{table*}

\subsection{LiDAR Semantic Segmentation}
\Cref{tab:semseg-stu,tab:semseg-kitti,tab:semseg-poss,tab:semseg-nuscenes} present detailed per-class LiDAR semantic segmentation results on STU inlier validation sequence and the proposed OoD datasets, comparing a standard semantic segmentation baseline with our approach that incorporates the losses described in~\cref{sec:methodology}. Our method achieves comparable performance with only a small degradation on nuScenes-OoD.
This drop is justified by the lower resolution and reduced number of points per scan in nuScenes, which may affect the construction of robust prototypes. The effect is further amplified by the severe class imbalance, where bicycle and motorcycle classes are present for only 0.01\% and 0.03\% of the data (approximately $10^5$ and $3\times 10^5$ points, respectively, out of the total), affecting contrastive learning, prototype building and the learned feature space. Consequently, these classes are often misclassified as manmade, which is significantly more present (15\%).

\section{Qualitative Results}

\begin{figure*}[t]
    \centering
    \includegraphics[width=0.8\linewidth]{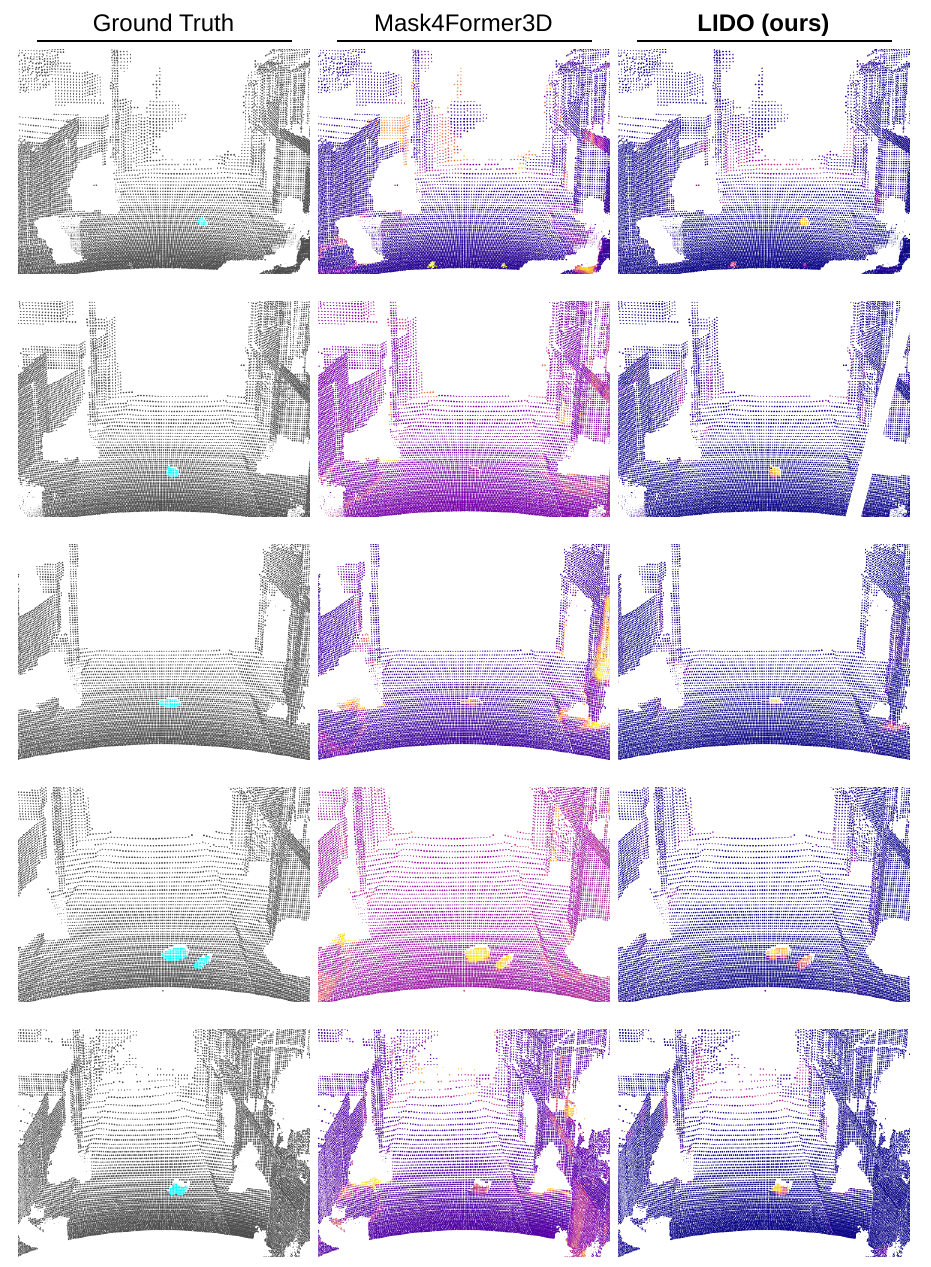}
    \caption{Qualitative comparison of 3D LiDAR anomaly segmentation results on STU validation set.}
    \label{fig:qualitative-stu}
\end{figure*}

\begin{figure*}[t]
    \centering
    \includegraphics[width=0.8\linewidth]{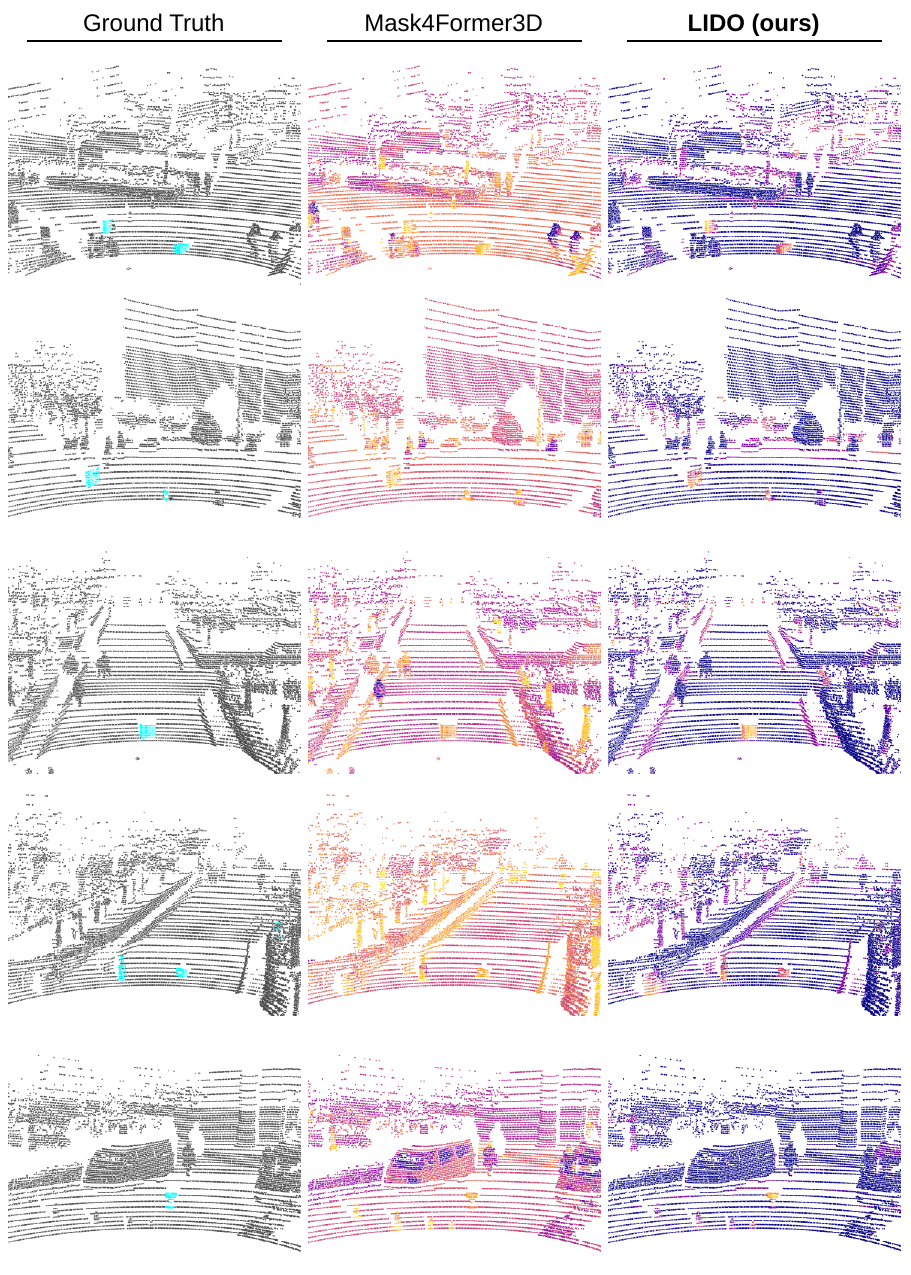}
    \caption{Qualitative comparison of 3D LiDAR anomaly segmentation results on SemanticPOSS-OoD \textit{Multi} split.}
    \label{fig:qualitative-poss}
\end{figure*}

\begin{figure*}[t]
    \centering
    \includegraphics[width=0.8\linewidth]{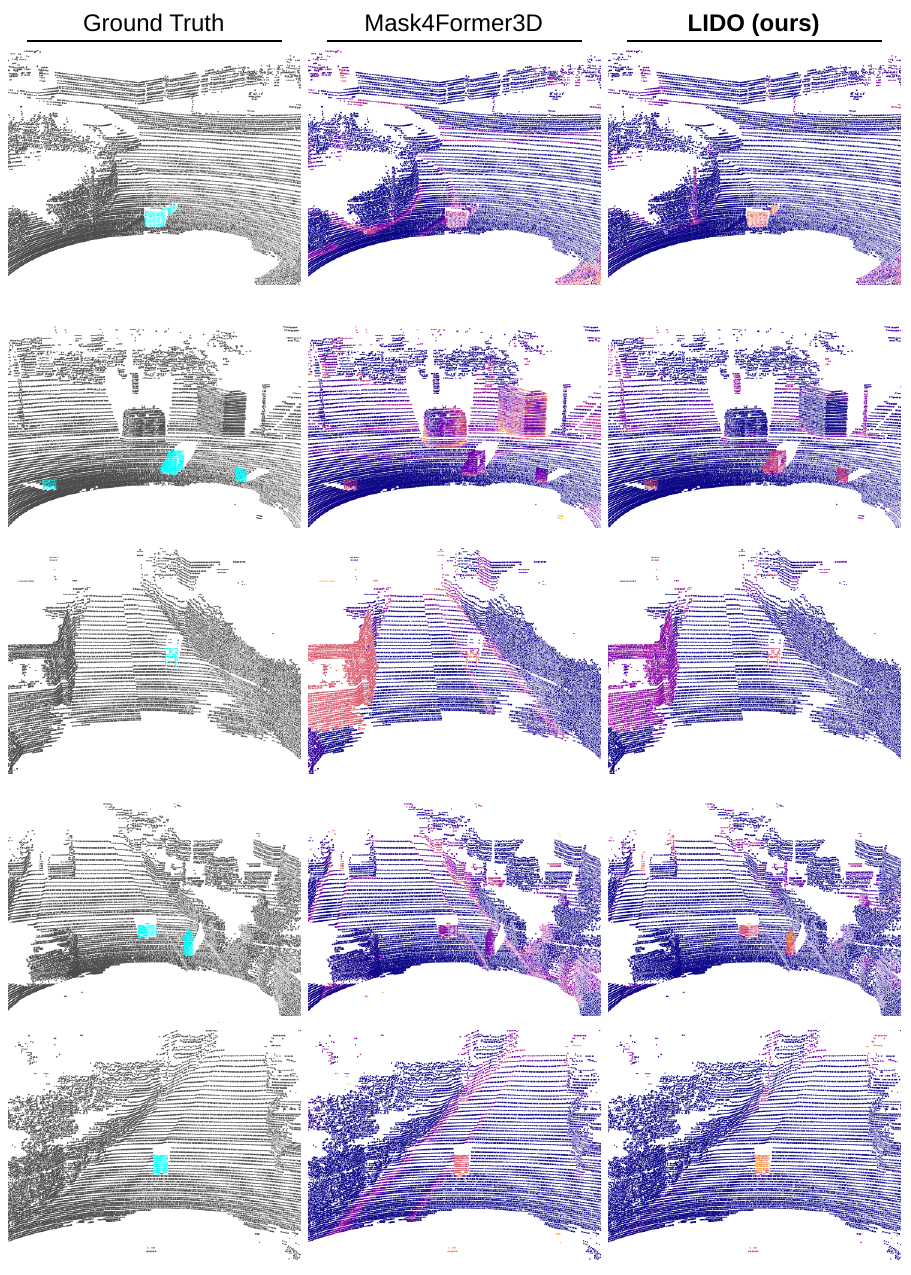}
    \caption{Qualitative comparison of 3D LiDAR anomaly segmentation results on SemanticKITTI-OoD \textit{Multi} split.}
    \label{fig:qualitative-kitti}
\end{figure*}

\begin{figure*}[t]
    \centering
    \includegraphics[width=0.8\linewidth]{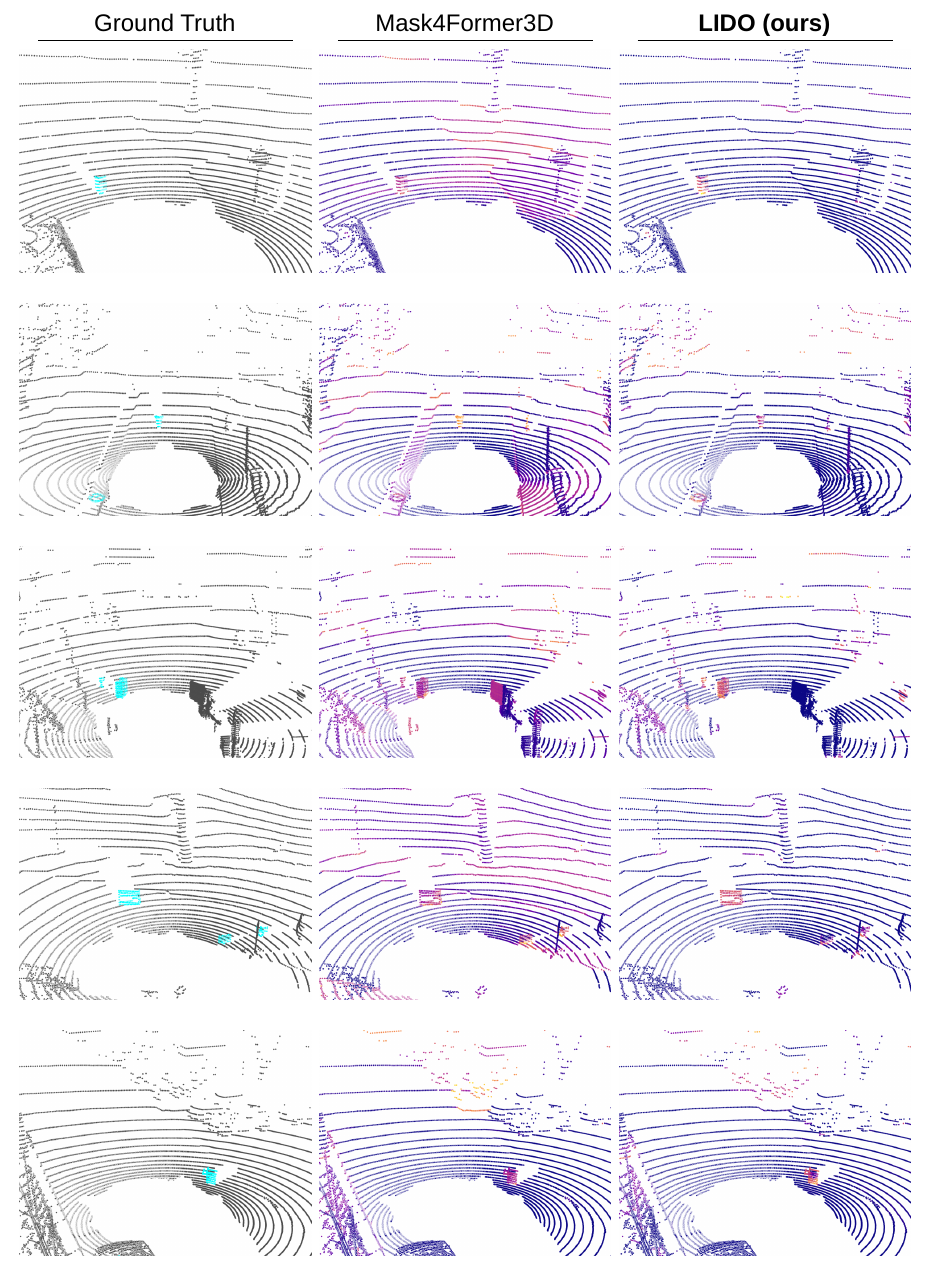}
    \caption{Qualitative comparison of 3D LiDAR anomaly segmentation results on nuScenes-OoD \textit{Multi} split.}
    \label{fig:qualitative-nuscenes}
\end{figure*}

We report further qualitative results for each dataset, comparing our method with the deep ensemble model in~\cite{nekrasov2025spotting}, built upon Mask4Former~\cite{yilmaz2024mask4former}.
~\Cref{fig:qualitative-stu,fig:qualitative-poss,fig:qualitative-kitti,fig:qualitative-nuscenes} show visualization of anomaly segmentation results on STU validation set, SemanticPOSS-OoD, SemanticKITTI-OoD and nuScenes-OoD datasets, respectively. We report both successful and failure cases. Our proposed approach demonstrates strong anomaly segmentation performance across the datasets, better or comparable to that of the deep ensemble model. For each figure, ground truth anomaly objects are shown in \textcolor{cyan}{cyan}, predicted anomaly scores follow the plasma color map from \textcolor{blue}{blue} to \textcolor{orange}{orange}, indicating increasing probability of a point being anomalous. Compared to the ensemble model used in~\cite{nekrasov2025spotting}, our approach produces fewer false positives. In the ensemble results, many inlier points that belong to the road and building classes are incorrectly marked in \textcolor{violet}{violet} or \textcolor{red}{red}, meaning that the model assigns them a high probability of being an anomaly. In contrast, our approach demonstrates more robust predictions on inlier classes as most points are colored in \textcolor{blue}{blue}, indicating low anomaly scores, with only limited uncertainty near class boundaries (e.g., road and sidewalk), within underrepresented classes or at long ranges, well-known challenges in standard semantic segmentation~\cite{kong2025calib3d}.


\end{document}